\documentclass[10pt,twocolumn,letterpaper]{article}
\usepackage[pagenumbers]{cvpr}

\graphicspath{{figures/}}
\usepackage{placeins}

\usepackage{pifont}
\newcommand{\benchyes}{\ding{51}}
\newcommand{\benchno}{\ding{55}}

\usepackage{colortbl}

\usepackage[most]{tcolorbox}
\definecolor{cvprblue}{rgb}{0.21,0.49,0.74}
\usepackage[breaklinks,colorlinks,allcolors=cvprblue]{hyperref}

\def\confName{CVPR}
\def\confYear{2027}
\title{HalluWord: A Benchmark for Word-Level Detection of Self-Generated Hallucinations in Vision-Language Models}
\author{Dengyao Shi$^{1,3}$\quad Tianyu Fu$^2$\quad Fei Hu$^2$\quad Anyang Su$^2$\quad Hanning Wang$^2$\\
Shuo Wang$^2$\quad Minghui Wu$^2$\quad Chenxu Zhao$^2$\quad Yang Yang$^1$\quad Zhen Lei$^{1,4,5,\dagger}$\\[3pt]
$^1$MAIS, Institute of Automation, Chinese Academy of Sciences\\
$^2$Mininglamp Technology\\
$^3$School of Advanced Interdisciplinary Sciences,\\
University of Chinese Academy of Sciences\\
$^4$CAIR, HKISI, Chinese Academy of Sciences\quad $^5$SCSE, FIE, M.U.S.T.\\
{\tt\small \{shidengyao2026,zhen.lei\}@ia.ac.cn}\quad
{\small $^{\dagger}$Corresponding author.}}

\hypersetup{pdfauthor={Dengyao Shi, Tianyu Fu, Fei Hu, Anyang Su, Hanning Wang, Shuo Wang, Minghui Wu, Chenxu Zhao, Yang Yang, Zhen Lei},
pdftitle={HalluWord: A Benchmark for Word-Level Detection of Self-Generated Hallucinations in Vision-Language Models}}

\begin{document}
\maketitle

\begin{abstract}
Differences in internal representations, tokenization, and native error
patterns hinder hallucination detector reuse across vision-language models
(VLMs). Training a separate detector for each model incurs additional
word-level annotation costs. We introduce HalluWord, a dataset of 91,869
usable native responses from six heterogeneous VLMs with word-level
hallucination annotations. HalluWord provides training data and an
evaluation benchmark of 1,500 shared image--question pairs from five
datasets, comprising 9,000 responses across ten question categories
with balanced AI-generated and non-AI-generated images. Responses remain
unmodified, preserving each model's naturally occurring hallucinations.
The benchmark supports both probe-based and VLM-based detection, with
per-model evaluation and protocols for assessing transfer without target
hallucination labels. We explore two cross-model approaches: OmniProbe
aligns heterogeneous representations to reuse a frozen source probe,
while OmniDetect uses a shared post-trained external VLM to localize
hallucinated words. HalluWord provides a common testbed for fine-grained
hallucination detection and cross-model detector reuse.
\url{https://github.com/Spell-shi/HalluWord}.
\end{abstract}

\section{Introduction}
\label{sec:intro}

\begin{figure*}[t]
    \centering
    \includegraphics[width=\textwidth]{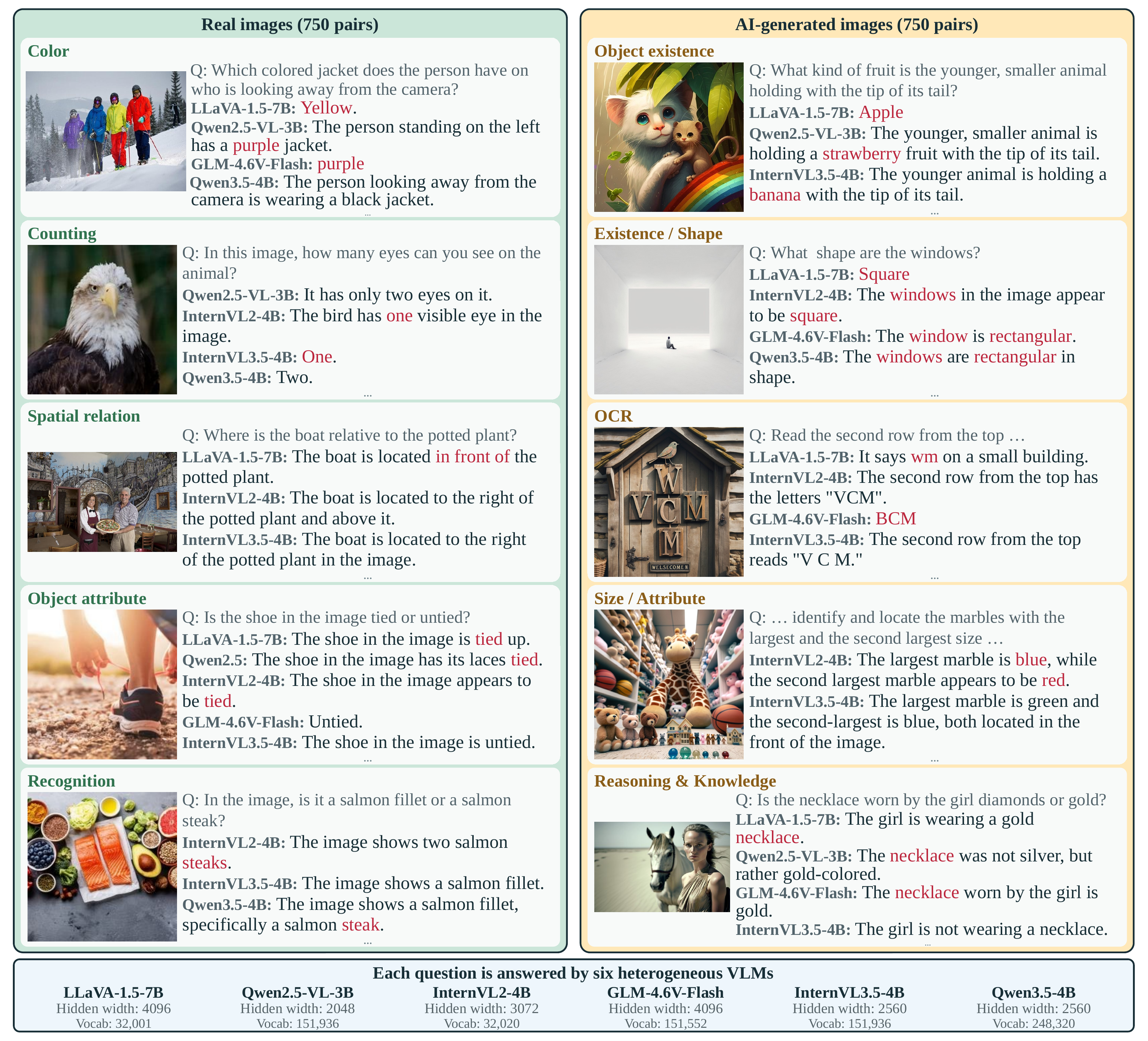}
    \caption{\textbf{HalluWord overview.} Each image--question pair has native
    responses from six heterogeneous VLMs, with word-level hallucination annotations.
    Selected responses are shown for real and AI-generated images;
    red words indicate annotated hallucinations.}
    \label{fig:halluword_teaser}
\end{figure*}

\begin{table*}[t]
    \centering
    \small
    \definecolor{benchgreen}{HTML}{198A48}
    \definecolor{benchred}{HTML}{CC403C}
    \definecolor{benchshade}{rgb}{0.941,0.918,0.969}
    \renewcommand{\benchyes}{\textcolor{benchgreen}{\ding{51}}}
    \renewcommand{\benchno}{\textcolor{benchred}{\ding{55}}}
    \setlength{\tabcolsep}{2.5pt}
    \renewcommand{\arraystretch}{1.25}
    \caption{Comparison of \textbf{hallucination localization benchmarks}.
    Samples counts evaluation responses, excluding training data.}
    \label{tab:benchmark_comparison}
    \newcommand{\benchstack}[1]{{\let\\\newline #1}}
    \dimen0=\dimexpr\textwidth-16\tabcolsep\relax
    \begin{tabular}{@{}>{\raggedright\arraybackslash}m{.164\dimen0}>{\centering\arraybackslash}m{.070\dimen0}>{\centering\arraybackslash}m{.136\dimen0}>{\centering\arraybackslash}m{.096\dimen0}>{\centering\arraybackslash}m{.091\dimen0}>{\centering\arraybackslash}m{.128\dimen0}>{\centering\arraybackslash}m{.107\dimen0}>{\centering\arraybackslash}m{.118\dimen0}>{\centering\arraybackslash}m{.09\dimen0}@{}}
        \toprule
        \textbf{Benchmark} & \textbf{Samples} &
        \benchstack{\bfseries Hallucination\\\bfseries origin} &
        \benchstack{\bfseries Annotation\\\bfseries unit} &
        \textbf{Categories} &
        \benchstack{\bfseries Image\\\bfseries sources} & \benchstack{\bfseries Evaluation\\\bfseries protocols} & \benchstack{\bfseries Per-VLM\\\bfseries results} & \benchstack{\bfseries OOD\\\bfseries evaluation} \\
        \midrule
        HalLoc~\cite{park2025halloc}\newline {\scriptsize CVPR 2025}
            & 15,154 & Injected & Token & 4 & Real & Probe-based & \benchno & \benchno \\
        MHALO~\cite{cai2025mhalo}\newline {\scriptsize Findings of ACL 2025}
            & 2,155 & \benchstack{Injected\\+ self-generated} & Token & 12 & Real & VLM-based & \benchno & \benchno \\
        HalluScope~\cite{jin2026halluscope}\newline {\scriptsize ACM MM 2026}
            & 733 & \benchstack{Injected\\+ self-generated} & Span & 12 & Real & VLM-based & \benchno & \benchno \\
        SHEEP~\cite{mickus2026sheep}\newline {\scriptsize EMNLP 2026}
            & 4,898 & \benchstack{Injected\\+ self-generated} & Span & 5 & Real & Both & \benchyes & \benchno \\
        \midrule

        \rowcolor{benchshade}
        \textbf{HalluWord (ours)}
            & \textbf{9,000} & \textbf{Self-generated} &
            \textbf{Word} & \textbf{10} & \benchstack{\textbf{Real}\\\textbf{+ AI-generated}} & \textbf{Both} & \benchyes & \benchyes \\
        \bottomrule
    \end{tabular}
    \vspace{3pt}
    \begin{minipage}{\textwidth}
        \footnotesize
        Both: probe- and VLM-based evaluation. Per-VLM results group
        native responses by their generating VLM; HalluWord probes each
        VLM's own states. OOD evaluates detection on responses from a VLM without using
        hallucination-labeled responses from that VLM for training.
        Injected includes human-written errors.
        \benchno: not reported.
    \end{minipage}
\end{table*}

Vision-language models (VLMs) can mix correct statements with
hallucinated details in the same response. Detecting that an answer is
unreliable is therefore insufficient: a detector must locate the words
expressing the error. Existing benchmarks evaluate hallucinations in
VLM outputs~\cite{rohrbach-etal-2018-object,li-etal-2023-evaluating,guan2024hallusionbench}
and, more recently, the localization of erroneous words or
spans~\cite{park2025halloc,cai2025mhalo,mickus2026sheep}.

Reusing a probe for word-level hallucination detection across VLMs is
challenging because the internal representations it relies on are not
directly compatible across
VLMs~\cite{park2025halloc,zhang2025dhcp,zhang2026vib,samuel2026uniprobe}.
Hidden-state dimensions can differ, and even equal-dimensional states
at the same layer index need not share compatible coordinates.
Different vocabularies and tokenizers further complicate correspondence
between response tokens. Training a separate supervised probe for each
new VLM requires hallucination annotations for that model's own responses,
repeating the labeling and training costs. These challenges motivate a
dataset and evaluation protocol for studying cross-model detection
without target-model hallucination labels.

Prior resources differ in response provenance and evaluation scope.
HalLoc and the constructed portion of MHALO use injected
errors~\cite{park2025halloc,cai2025mhalo}, whose distribution may differ
from models' own mistakes. M-HalDetect provides annotated model
responses~\cite{gunjal2024mhaldetect}; SHEEP includes outputs from five
VLMs and human-written responses~\cite{mickus2026sheep}; and VisionHall
collects descriptions from twelve models~\cite{wada2026zina}.
We build on these resources by focusing on native answers to shared
questions, with training data and protocols for testing detector reuse
across VLMs, as summarized in \cref{tab:benchmark_comparison}.

We introduce HalluWord, a benchmark of 9,000 native responses from six
heterogeneous VLMs to 1,500 shared image--question pairs, accompanied
by training and validation data. It covers ten question categories and
both AI-generated and non-AI-generated images. Word-level annotation
combines two-model agreement with human review of disagreements.
The shared-question design supports comparisons across generating
models using both internal-state probes and external VLM detectors
(\cref{fig:halluword_teaser,sec:benchmark}).

Using HalluWord, we explore two approaches to reducing the need for
model-specific probe training and hallucination annotations, motivated
by cross-model detection
research~\cite{chuang-etal-2024-lookback,gupta2026external}.
OmniProbe aligns target representations to a frozen source probe using
unlabeled responses and word correspondence. OmniDetect uses a shared
post-trained external VLM to identify hallucinated words across response
models. Both target reuse without hallucination labels for responses
generated by target VLMs; \cref{fig:cross_model_detection,sec:method}
describes the two approaches.

Our contributions are threefold:
\begin{itemize}
    \item \textbf{HalluWord dataset.} We provide training data and a
    shared-question benchmark with word-level hallucination annotations
    on native responses from six heterogeneous VLMs.
    \item \textbf{Unified detection evaluation.} We compare training-free,
    probe-based, and VLM-based methods on responses generated by different
    VLMs, and assess whether a detector can be reused across VLMs without
    hallucination annotations for the target model's responses.
    \item \textbf{Cross-model detection approaches.} OmniProbe reuses
    a source probe through representation alignment; OmniDetect uses
    a shared external VLM for word-level detection.
\end{itemize}

\section{Related Work}
\label{sec:related_work}

\paragraph{Hallucination benchmarks.}
CHAIR, POPE, and AMBER evaluate object hallucinations and related visual
errors~\cite{rohrbach-etal-2018-object,li-etal-2023-evaluating,wang2023amber},
while HallusionBench, HaloQuest, and VHTest examine visual reasoning and
hallucination failures~\cite{guan2024hallusionbench,wang2024haloquest,huang-etal-2024-visual}.
These benchmarks primarily assess the responding model's reliability.
For detection within existing responses, M-HalDetect and MHaluBench
provide sub-sentence annotations or segment/claim verification
settings~\cite{gunjal2024mhaldetect,chen2024mhalubench}.
HalLoc uses injected errors for token-level localization, and MHALO
combines constructed and native-response data~\cite{park2025halloc,cai2025mhalo}.
SHEEP and SHROOM-Visions study span detection across languages and
models, including human-written evaluation
examples~\cite{mickus2026sheep,vazquez2026shroom}; VisionHall provides
annotated native descriptions from multiple models~\cite{wada2026zina}.
HalluWord collects native responses from different VLMs, provides
word-level hallucination annotations, and supports both probe-based
and VLM-based evaluation of hallucination detection.

\paragraph{Hallucination detection and transfer.}
Internal-signal methods exploit token statistics, attention, hidden
representations, and visual--textual grounding, as explored by MetaToken,
DHCP, HalLocalizer, VIB-Probe, HalluShift++, patch-level grounding, and
UniProbe~\cite{fieback2025metatoken,zhang2025dhcp,park2025halloc,zhang2026vib,nath2025hallushift,nguyen2026tokengrounding,samuel2026uniprobe}.
HALP instead predicts hallucination risk before response
generation~\cite{kogilathota-etal-2026-halp}.
External evaluators verify or localize errors through factual checks,
tools, or adapted VLMs, including FaithScore, UniHD, HaloDet, ZINA,
HalluScope, and SHROOM-Visions
systems~\cite{jing-etal-2024-faithscore,chen2024mhalubench,cai2025mhalo,wada2026zina,jin2026halluscope,branescu-dascalu-2026-team,jara-mikolajczak-etal-2026-team,bubus-2026-team}.

Related work studies word-level annotation, cross-model transfer, and
observer-based detection in text~\cite{niu-etal-2024-ragtruth,chuang-etal-2024-lookback,gupta2026external},
as well as adaptation to self-generated VLM responses~\cite{samuel2026uniprobe}.
ComnHallu aligns principal-component subspaces for cross-model
object-hallucination detection within TruthPrInt~\cite{duan2025truthprint}.
We instead use word-corresponding native responses to learn mappings
into a frozen source probe's space.
We study cross-model word-level hallucination detection without requiring
hallucination labels from target VLMs, through probe-based and
VLM-based approaches.

\section{The HalluWord Dataset}
\label{sec:benchmark}
\begin{figure*}[t]
    \centering
    \includegraphics[width=\textwidth]{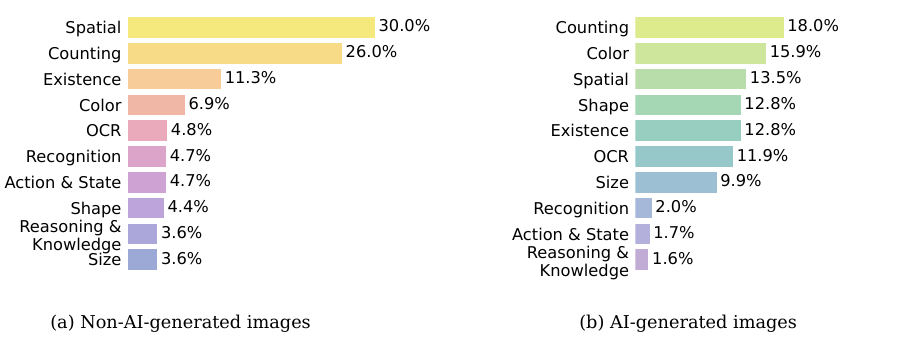}
    \caption{Question-category distributions in the HalluWord benchmark.
    (a) Non-AI-generated images. (b) AI-generated images. Each group contains
    750 image--question pairs; percentages are calculated within each group.
    Each question has native responses from six VLMs.}
    \label{fig:category_distribution}
\end{figure*}

HalluWord provides training data for learning hallucination detectors and
an evaluation benchmark for measuring word-level detection across
heterogeneous VLMs. Both use native model responses with
word-level annotations. We describe their collection and annotation below.

\paragraph{Data collection.}
We collect existing image--question pairs from five datasets:
A-OKVQA~\cite{schwenk2022aokvqa}, MMVP~\cite{tong2024eyes},
CV-Bench from Cambrian-1~\cite{tong2024cambrian},
HaloQuest~\cite{wang2024haloquest}, and
VHTest~\cite{huang-etal-2024-visual}.
Six VLMs with heterogeneous architectures generate native responses to
these inputs. For each image--question pair, each VLM samples one response
with temperature $1.0$, top-$p=0.95$, and top-$k=20$.

The responses
are retained without editing them to insert hallucinations. Annotation
identifies words expressing content that contradicts or is unsupported
by the visual evidence; it does not alter the response text.

\paragraph{Data annotation pipeline.}
We first use \texttt{gpt-6.1-sol} to check reference ground-truth (GT)
answers for ambiguity and distinguish genuine hallucinations from
image-supported responses that differ from the reference. Such valid
responses are retained with non-hallucination word labels.

We then obtain word-level annotations independently from
\texttt{gpt-6.1-sol} and \texttt{claude-5.0-opus}, using the same prompt
template with the image, question, native response, and reference GT.
Across training and benchmark responses from all six VLMs, 83.34\%
receive identical word-label sets, which we retain. For the 8.23\% with
partially overlapping annotations, we retain the intersection and conduct
targeted human review of candidate hallucinated words outside it.
The remaining 8.43\% undergo human review. Percentages are measured at
the response level, and the original response text remains unchanged.
An additional human audit, independent of the two annotation models,
examined 300 sampled responses with overlapping hallucination annotations
across all five datasets and identified five associated questions as
unsuitable for word-level hallucination detection.

\paragraph{Dataset statistics.}
Across the five sources, we collect 15,341 image--question pairs and
92,046 native responses from six heterogeneous VLMs.
Following manual review, 177 responses, accounting for only 0.19\% of
all collected responses, were excluded because they contained extended
passages whose hallucination could not be reliably localized to individual
words. These responses were unsuitable for word-level annotation and
were removed before dataset splitting, leaving 91,869 usable responses
with 828,962 labeled words.
We select 1,500 pairs to form the evaluation benchmark, balanced equally
between AI-generated and non-AI-generated images (750 pairs each).
Keeping all six VLMs' responses for each selected pair yields
9,000 evaluation responses with 87,471 labeled words. The remaining
13,841 pairs form the training and validation pool, comprising 82,869 usable responses
with 741,491 labeled words. These word counts include both hallucinated
and non-hallucinated words retained after label and alignment filtering.
Our evaluation protocol reserves 500 image--question pairs from this
pool for validation, with all six VLM responses per pair (3,000 responses),
leaving 13,341 pairs for training. Splitting is performed at the question
level: all responses to a question remain in one split, and the training,
validation, and benchmark splits are mutually disjoint.
Validation and benchmark questions are excluded from detector fitting,
PCA fitting, and representation-alignment fitting. Supervised training uses labels only
from VLMs permitted by the evaluation protocol; the held-out
VLM's hallucination labels are excluded.

We organize the benchmark questions into ten categories:
Counting, Spatial, Existence, Color, Shape, OCR, Size, Recognition,
Action \& State, and Reasoning \& Knowledge.
\Cref{fig:category_distribution} shows their distribution.
Across all six VLMs, 56.07\% of responses to non-AI-generated images
and 58.04\% of responses to AI-generated images contain at least one
hallucinated word.
Spatial covers relative position, orientation, viewpoint, and occlusion;
Recognition concerns object or category identity, while Action \& State
covers activities, poses, and conditions such as open versus closed or
tied versus untied. Counts refer to image--question pairs, each associated
with six model responses.

\paragraph{Evaluation protocol.}
Given an image, a question, and an existing response, a detector assigns
a hallucination score to each response word. Question categories describe
input coverage; they are not word-level hallucination-type labels.
The prediction target is binary hallucination detection at the word level.
For the main comparison in \cref{tab:detection_main}, we reserve
Qwen3.5-4B as the held-out VLM, excluding its
hallucination labels from detector training, model selection, and
threshold selection. OmniProbe may use unlabeled target training
responses and their internal states for representation alignment;
all benchmark questions are excluded from alignment fitting.
We report word-level detection metrics separately for each of the
six VLMs, highlighting the held-out VLM's results.
A separate rotating-source analysis in \cref{sec:cross_model_transfer}
tests whether each VLM can supply a reusable probe.
For probe-based methods, evaluation uses each VLM's own internal
states, exposing differences in representation dimensions and tokenization.
This protocol tests VLM shift on shared image--question inputs.

\section{From Model-Specific Detection to Cross-Model Detection}
\label{sec:method}

\begin{figure*}[t]
    \centering
    \includegraphics[width=\textwidth]{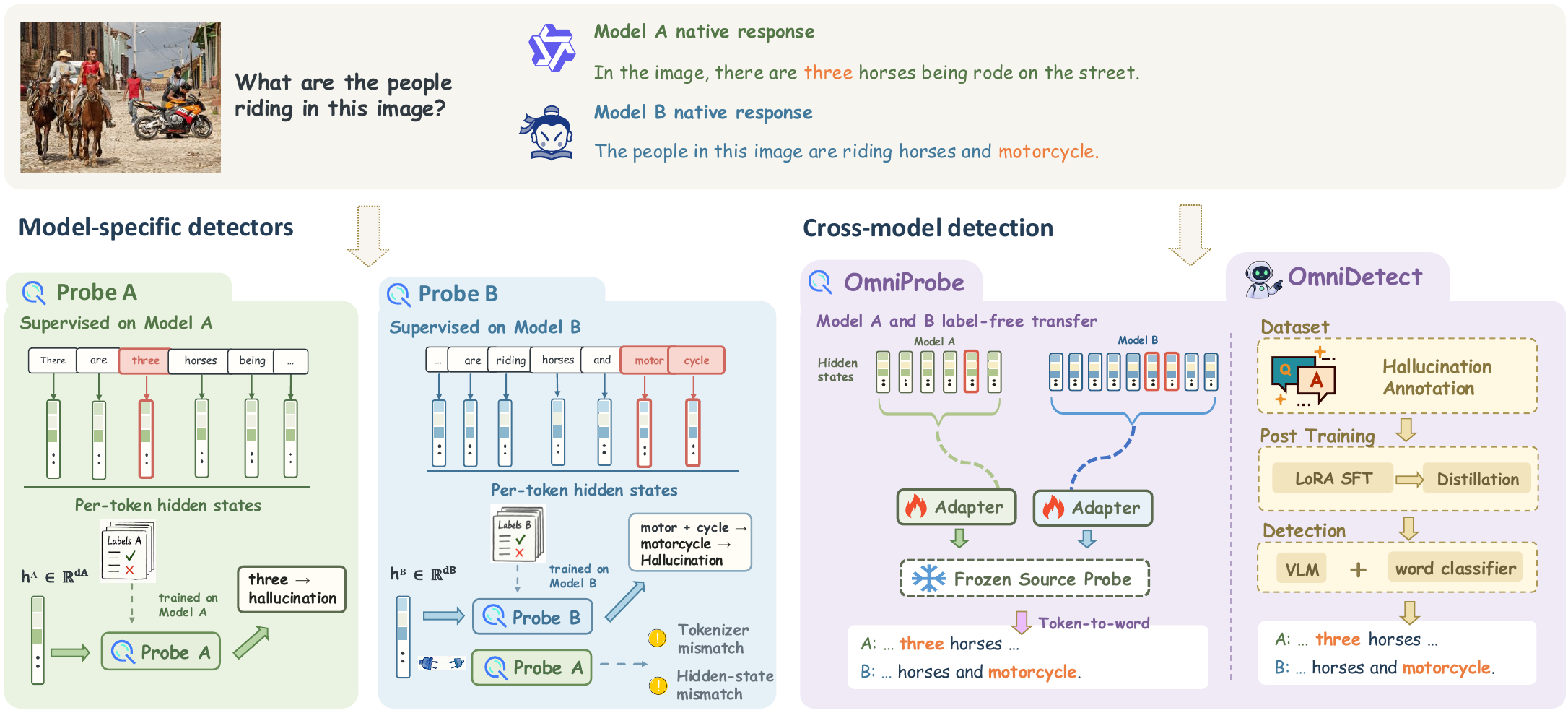}
    \caption{\textbf{Cross-model word-level hallucination detection.}
    Left: model-specific probes. Right: OmniProbe and OmniDetect,
    two independent routes to detector reuse.
    Highlighted words illustrate hallucination localization.}
    \label{fig:cross_model_detection}
\end{figure*}

We study two routes to cross-model word-level hallucination detection
(\cref{fig:cross_model_detection}). OmniProbe reuses a frozen source probe
through representation alignment, while OmniDetect applies a shared
external VLM to responses from different generators.

\subsection{OmniProbe: Reusing a Frozen Source Probe}
\label{sec:omniprobe}

\paragraph{Source probe and paired representations.}
We train a probe $f_{\theta_s}$ on source-VLM states and binary
hallucination labels, then freeze both the probe and its source-fitted
projection. We replay each target response through the source VLM and
match words by their character spans in the unchanged text, establishing
correspondence across different tokenizations. The source VLM remains
frozen and receives the same image and question.
Let $u_w^t$ and $z_w^{s\leftarrow t}$
denote the paired target and projected source word representations.
Neither reference answers nor target hallucination labels are required.

\paragraph{Cross-model alignment.}
For each target VLM, we learn a map
$A_t:\mathbb R^{d_t}\rightarrow\mathbb R^k$ into the frozen source
space using Ridge regression.
Let $\mu_t$ and $\mu_s$ be the means of paired target and projected
source word representations in the alignment training split. We fit
\begin{equation}
 W^*=\arg\min_W\sum_w
 \|W(u_w^t-\mu_t)-(z_w^{s\leftarrow t}-\mu_s)\|_2^2
 +\lambda\|W\|_F^2,
 \label{eq:ridge_alignment}
\end{equation}
and define $A_t(h)=A_{\mathrm R}(h)=W^*(h-\mu_t)+\mu_s$.
This objective matches paired features without target hallucination labels.

\paragraph{Probe reuse.}
At inference, we map each target token state into the source space and
apply the frozen probe to the resulting sequence. Word-level scores are
obtained by pooling the corresponding token probabilities:
\begin{equation}
 p_w^t=\operatorname{Pool}_{i\in\mathcal T_t(w)}
       [f_{\theta_s}(A_t(H^t))]_{i,1},
 \label{eq:omniprobe_word_prediction}
\end{equation}
where $H^t$ is the target response's state sequence,
$\mathcal T_t(w)$ contains tokens corresponding to word $w$, and class
$1$ denotes hallucination. The source decision threshold is reused
across targets. Source-model replay is needed only to construct alignment
pairs; detection uses the target VLM's own states and the frozen
source probe.

\subsection{OmniDetect: A Shared VLM Detector}
\label{sec:omnidetect}

OmniDetect evaluates responses from different VLMs with one
external VLM. Its input $x=(I,Q,R)$ contains an image, a question, and
an existing native response $R=(w_1,\ldots,w_n)$. It predicts a binary
hallucination probability for each word, using the evaluator's own
representations. The generating VLM's hidden-state dimension and
vocabulary therefore do not determine the detector architecture.

\paragraph{Stage 1: word-level supervised fine-tuning.}
Given $x=(I,Q,R)$, OmniDetect assigns each response word $w_i$ a
hallucination probability $p_{\psi,i}(x)\in[0,1]$ through a binary
classification head. We fine-tune the detector using word-level labels
$y_i\in\{0,1\}$, where $y_i=1$ indicates hallucination, with the objective
\begin{equation}
    \mathcal L_{\mathrm{SFT}}
    =\frac{1}{N_y}\sum_{(x,i)\in\mathcal S}
      \operatorname{BCE}\!\left(p_{\psi,i}(x),y_i\right),
    \label{eq:sft}
\end{equation}
where $\mathcal S$ contains response words with valid annotations,
$N_y=|\mathcal S|$, and BCE denotes binary cross-entropy.

\paragraph{Stage 2: word-level distillation.}
We initialize a frozen teacher and a trainable student from the SFT
checkpoint. The teacher receives $(I,Q,R,r^\star)$, where $r^\star$ is
the reference answer; the student receives only $(I,Q,R)$.
The reference answer precedes the response, making it accessible to
response-token representations under causal masking.
Both predict a binary hallucination distribution for each word in the
same unchanged response.
Let $\mathbf q_i$ and $\mathbf p_i$ denote the teacher and student
distributions for word $i$, respectively. We align corresponding response
words and train the student to match the teacher's soft predictions:
\begin{equation}
    \mathcal L_{\mathrm{distill}}
    = \frac{1}{N_d}\sum_{(x,i)\in\mathcal A}
      D_{\mathrm{KL}}(\mathbf q_i\,\|\,\mathbf p_i),
    \label{eq:distillation}
\end{equation}
where $\mathcal A$ contains valid aligned response words across the
distillation data and $N_d=|\mathcal A|$. This objective transfers word-level
hallucination judgments rather than vocabulary-generation distributions.

\paragraph{Inference.}
At inference, a shared student and decision threshold detect hallucinated
words across VLMs using only $(I,Q,R)$, without reference answers.

\section{Experiments}
\label{sec:experiments}

\begin{table*}[t]
\centering
\small
\setlength{\tabcolsep}{3pt}
\renewcommand{\arraystretch}{1.15}
\caption{\textbf{Comparison of hallucination detection methods on HalluWord.}
Columns correspond to VLMs. Entries report $F_1$
(\%); the benchmark contains 1,500 responses per VLM.
Bold and underlined values indicate the best and second-best results in each column.}
\label{tab:detection_main}
\definecolor{ourslavender}{rgb}{0.941,0.918,0.969}
\dimen0=\dimexpr\textwidth-14\tabcolsep\relax
\begin{tabular}{@{}>{\raggedright\arraybackslash}m{.31\dimen0}*{7}{>{\centering\arraybackslash}m{.09857142\dimen0}}@{}}
\toprule
& \multicolumn{6}{c}{In-domain (ID)} & \multicolumn{1}{c}{OOD} \\
\cmidrule(lr){2-7}\cmidrule(l){8-8}
Method & \shortstack{LLaVA\\1.5-7B} & \shortstack{Qwen2.5\\VL-3B} &
\shortstack{InternVL2\\4B} & \shortstack{GLM-4.6V\\Flash} &
\shortstack{InternVL3.5\\4B} & \shortstack{Avg.\\(ID)} &
\shortstack{Qwen3.5\\4B} \\
\midrule
\multicolumn{8}{l}{\textit{Training-free uncertainty scores}} \\
Token NLL~\cite{fieback2025metatoken} & 35.37 & 27.85 & 26.37 & 16.59 & 22.89 & 25.81 & 22.79 \\
Token Entropy~\cite{fieback2025metatoken} & 32.49 & 28.10 & 24.14 & 16.69 & 21.89 & 24.66 & 21.47 \\
Max Probability~\cite{hendrycks2017baseline} & 33.30 & 26.19 & 22.92 & 17.57 & 21.54 & 24.30 & 19.21 \\
\midrule
\multicolumn{8}{l}{\textit{Probe-based methods}} \\
Linear Probe$^{\dagger}$ & 67.99 & 53.30 & 57.34 & 50.51 & 51.02 & 56.03 & -- \\
MM$^{\dagger}$~\cite{marks2024geometry} & 57.75 & 33.75 & 30.63 & 23.84 & 24.69 & 34.13 & -- \\
SAPLMA-Token$^{\dagger}$~\cite{azaria2023saplma} & 68.43 & 55.22 & 57.29 & 52.43 & 54.09 & 57.49 & -- \\
MLP Probe$^{\dagger}$ & 68.57 & 55.10 & 56.97 & 52.52 & 53.94 & 57.42 & -- \\
HalLocalizer$^{\dagger}$~\cite{park2025halloc} & 65.99 & 50.03 & 53.23 & 47.10 & 50.04 & 53.28 & -- \\
BiGRU Probe$^{\dagger}$ & 68.69 & 55.18 & 58.04 & 54.07 & 53.12 & 57.82 & -- \\
CPS-Conv1D$^{\dagger}$~\cite{srivastava2026cps} & 68.33 & 55.45 & 57.16 & 51.40 & 53.82 & 57.23 & -- \\
Lyapunov Probe$^{\dagger}$~\cite{luan2026lyapunov} & 68.51 & 55.83 & 57.88 & 52.90 & 53.96 & 57.82 & -- \\
\rowcolor{ourslavender}
\textbf{OmniProbe (ours)}$^{\ddagger}$ & 67.73 & 52.97 & 55.58 & 42.06 & 47.20 & -- & 47.81 \\
\midrule
\multicolumn{8}{l}{\textit{VLM-based methods}} \\
GPT-6.1-sol & \underline{78.70} & \textbf{69.82} & \underline{70.17} & 63.28 & \textbf{70.67} & \textbf{70.53} & \textbf{65.84} \\
Gemini-3.8-Flash & 78.32 & 68.12 & 67.07 & \textbf{65.17} & \underline{68.39} & 69.41 & 63.88 \\
DeepSeek-V4.1-Flash & 72.20 & 61.06 & 60.35 & 56.09 & 61.48 & 62.23 & 55.75 \\
GLM-5.3-Flash & 62.45 & 50.99 & 51.95 & 46.52 & 53.34 & 53.05 & 48.59 \\
Qwen2.5-VL-72B~\cite{bai2025qwen25vl} & 62.36 & 49.19 & 48.79 & 37.29 & 44.48 & 48.42 & 40.60 \\
Qwen3.8-27B~\cite{qwen2026qwen38} & 67.97 & 58.33 & 58.62 & 47.27 & 55.77 & 57.59 & 52.37 \\
Qwen3.6-35B-A3B~\cite{qwen2026qwen36} & 57.09 & 44.82 & 43.71 & 30.08 & 39.06 & 42.95 & 33.02 \\
HalluScope-8B~\cite{jin2026halluscope} & 53.51 & 42.07 & 42.37 & 30.99 & 39.75 & 41.74 & 35.42 \\
\rowcolor{ourslavender}

\textbf{OmniDetect-27B (ours)} & \textbf{78.93} & \underline{68.87} & \textbf{70.21} & \underline{64.48} & 67.96 & \underline{70.09} & \underline{64.52} \\
\bottomrule
\end{tabular}
\vspace{3pt}
\begin{minipage}{\textwidth}\footnotesize
$\dagger$ \textbf{Per-VLM supervision:} A separate probe is trained and selected for each VLM
using hallucination labels for that VLM's own responses.
\par
$\ddagger$ \textbf{Source-only supervision:} OmniProbe uses hallucination labels only from LLaVA,
making LLaVA its sole ID VLM and all other VLMs OOD transfer targets; alignment uses unlabeled responses.
-- denotes not applicable under the current evaluation protocol.
\end{minipage}
\end{table*}

\subsection{Experimental Setup}
\label{sec:experimental_setup}

\paragraph{Evaluation protocol.}
We follow the disjoint splits defined in \cref{sec:benchmark}.
ID/OOD status depends on the
VLMs whose hallucination labels are used to train each detector.
In the main comparison, LLaVA is the sole ID VLM for OmniProbe; the other five VLMs are OOD
transfer targets, with alignment learned from unlabeled training responses.
OmniDetect results use five in-domain VLMs for
SFT and distillation. Supervised probe checkpoints
and thresholds are selected exclusively by word-level F1 on the separate
validation split using labels from the supervising VLM, then fixed before
benchmark evaluation. Target hallucination labels are unavailable during
transfer. Uncertainty-score thresholds are calibrated on the LLaVA
validation responses and fixed across VLMs.

\paragraph{Hidden-state extraction.}
All probe-based methods use response-token hidden states from the
midpoint of each VLM's language decoder: the output of decoder block
$L/2$ (one-based indexing), where $L$ is the number of decoder blocks.

\paragraph{Prompted VLM baselines.}
The prompted VLM baselines in \cref{tab:detection_main} and the zero-shot
\textit{w/o SFT} baseline in \cref{tab:omnidetect_ablation} use the same
fixed instruction and input template across all six VLMs,
without VLM-specific prompt tuning. The template takes the image,
original question, native response, and numbered response words, and
returns hallucinated word IDs. The full template is provided in \cref{sec:supp_shared_prompt}.

\paragraph{OmniProbe.}
We train an MLP--BiGRU probe on LLaVA-1.5-7B states, reduced from
4,096 to 256 dimensions by source-fitted PCA, using class-weighted
cross-entropy. Tokens are matched to response words by character offsets; their
hallucination probabilities are averaged within each word. The checkpoint
and threshold are selected by word-level F1 on the LLaVA responses in the separate validation split
and reused across target models.
For the alignment ablation, Ridge and Probe-guided Alignment share
the frozen source probe; ComnHallu uses the same probe architecture
retrained on its normalized source projections.

\paragraph{OmniDetect.}
We use Qwen3.8-27B with a linear word-level binary classification head.
LoRA~\cite{hu2022lora} is applied to the language decoder with rank 16
and $\alpha=32$. Only the LoRA parameters and classification head are
trained; the pretrained backbone and vision encoder remain frozen.
SFT and distillation use the same training responses, with word-level
annotations providing SFT supervision and reference-conditioned teacher
predictions providing distillation targets.

\subsection{Evaluation Metrics}
\label{sec:metrics}

Our main metric is standard word-level $F_1$, treating hallucinated words
as the positive class. A predicted word is correct only if its gold label
is positive. We pool true positives (TP), false positives (FP), and false
negatives (FN) over the evaluated responses from each VLM and compute
\begin{equation}
    F_1=\frac{2\mathrm{TP}}{2\mathrm{TP}+\mathrm{FP}+\mathrm{FN}}.
    \label{eq:word_f1}
\end{equation}
Averages across VLMs are unweighted means of their individual scores.
For methods that produce continuous scores, we additionally report AUROC
and average precision (AP) in \cref{sec:supp_detection_metrics}.

\subsection{Comparison with Existing Methods}
\label{sec:method_comparison}

\Cref{tab:detection_main} compares training-free uncertainty scores,
supervised probes, and VLM-based detectors on HalluWord,
reporting $F_1$ separately for each VLM. Lyapunov
Probe~\cite{luan2026lyapunov} adapts derivative-based regularization to
our word-level detection setting using cached token features. Conventional
probes use per-VLM supervision; OmniProbe uses LLaVA supervision
and transfers to the other five VLMs. The table's ID/OOD column grouping
follows the five-VLM protocol, with Qwen3.5-4B held out.
Avg. (ID) is the unweighted mean over the five ID columns and is omitted
for OmniProbe, whose ID evaluation is limited to LLaVA.
\subsection{Analysis by Question Category}
\label{sec:category_analysis}

\begin{figure}[t]
    \centering
    \includegraphics[width=\columnwidth]{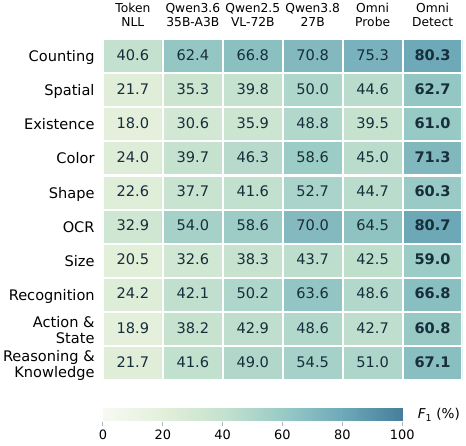}
    \caption{\textbf{Hallucination detection across question categories.}
    Entries report $F_1$ (\%), computed within each category and
    VLM and then averaged equally across six VLMs.
    Bold marks the highest score in each row; Qwen baselines use zero-shot prompting.}
    \label{fig:category_detection_heatmap}
\end{figure}

\Cref{fig:category_detection_heatmap} shows that Counting and OCR yield
the two highest category scores for every method. Relative strengths
vary elsewhere: OmniProbe outperforms zero-shot Qwen3.8-27B on Counting
but trails it on Recognition and Color. OmniDetect achieves the highest
score in all ten categories, reaching 80.7 on OCR and 80.3 on Counting.
Size and Shape remain its lowest-scoring categories (59.0 and 60.3).

\subsection{Cross-Model Transfer}
\label{sec:cross_model_transfer}

\begin{figure}[t]
    \centering
    \includegraphics[width=\columnwidth]{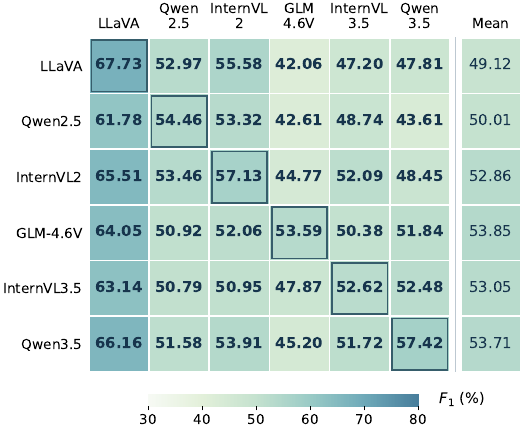}
    \caption{\textbf{OmniProbe cross-model transfer.}
    Word-level $F_1$ (\%); rows are sources and columns are targets.
    Outlined cells denote within-model evaluation and are excluded from Mean.}
    \label{fig:omniprobe_transfer}
\end{figure}

\Cref{fig:omniprobe_transfer} evaluates all 30 directed transfers among
six architecturally heterogeneous VLMs under a separate rotating-source
protocol. Each row uses only its source VLM's hallucination labels for
probe training; the other five VLMs provide unlabeled
alignment data. Thus, Qwen3.5-4B supplies supervision only in its own
source row, while remaining a label-free target in all other rows.
This analysis tests unsupervised transfer of a supervised source probe;
it does not change the fixed held-out setting of the main comparison.
In every direction, Ridge alignment reuses the frozen source probe
without target-model hallucination labels or probe retraining. Mean transfer $F_1$ across the five targets
ranges from 49.12 to 53.85. These results demonstrate probe reuse across
all six source models despite differences in hidden-state dimensions
and tokenization, enabling word-level hallucination detection on new
response generators through unlabeled representation alignment.

\subsection{Ablation Studies}
\label{sec:ablations}

\subsubsection{OmniProbe Alignment Strategies}
\label{sec:alignment_comparison}

Probe-guided Alignment adds a nonlinear correction trained to match
the frozen source probe's word-level soft predictions without target
labels. It provides no consistent improvement over Ridge
(\cref{tab:omniprobe_alignment}).
We include ComnHallu as a subspace-alignment baseline.
We therefore retain Ridge as the default alignment.

\begin{table}[!htbp]
    \centering
    \small
    \setlength{\tabcolsep}{2pt}
    \renewcommand{\arraystretch}{1.12}
    \caption{\textbf{OmniProbe alignment comparison.}
    AUROC (\%); LLaVA is the source.
    N/A: incompatible hidden-state dimensions.}
    \label{tab:omniprobe_alignment}
    \begin{tabular*}{\columnwidth}{@{\extracolsep{\fill}}lccc@{}}
        \toprule
        Target VLM & ComnHallu & Probe-guided & Ridge \\
        & \cite{duan2025truthprint} & Alignment & \\
        \midrule
        Qwen2.5-VL-3B & N/A & \textbf{91.03} & 90.71 \\
        InternVL2-4B & N/A & \textbf{93.03} & 92.95 \\
        GLM-4.6V-Flash & 51.62 & 89.66 & \textbf{89.90} \\
        InternVL3.5-4B & N/A & 92.07 & \textbf{92.08} \\
        Qwen3.5-4B & N/A & 90.95 & \textbf{91.09} \\
\midrule
        Mean  & -- & \textbf{91.35} & \textbf{91.35} \\
        \bottomrule
    \end{tabular*}
\end{table}

\subsubsection{OmniDetect Post-training Components}
\label{sec:omnidetect_ablations}

\Cref{tab:omnidetect_ablation} compares the training components of OmniDetect, which undergoes one epoch of SFT followed by one epoch of distillation. The SFT-only baseline uses two epochs of supervised training with the same number of student updates. “+ reference” provides OmniDetect with the ground-truth answer to the question during detection.
\definecolor{ourslavender}{rgb}{0.941,0.918,0.969}
\par\medskip\noindent
\begin{minipage}{\columnwidth}
    \centering
    \captionsetup{hypcap=false}
    \footnotesize
    \setlength{\tabcolsep}{2pt}
    \renewcommand{\arraystretch}{1.15}
    \captionof{table}{\textbf{Component ablations of OmniDetect.}
    Scores are percentages.}
    \label{tab:omnidetect_ablation}

    \dimen0=\dimexpr\columnwidth-10\tabcolsep\relax
    \begin{tabular}{>{\raggedright\arraybackslash}m{.22\dimen0}*{4}{>{\centering\arraybackslash}m{.195\dimen0}}}
        \toprule
        Setting & \shortstack{$F_1$\\ID} & \shortstack{AUROC\\ID} &
        \shortstack{AP\\ID} & \shortstack{$F_1$\\OOD} \\
        \midrule
        w/o SFT & 57.59 & -- & -- & 52.37 \\
        w/o KD & 68.12 & 95.36 & 74.23 & 63.45 \\
        \rowcolor{ourslavender}
        \textbf{OmniDetect} & $70.09_{\pm 0.16}$ & $96.03_{\pm 0.03}$ &
        $75.43_{\pm 0.11}$ & $64.52_{\pm 0.26}$ \\
        \rowcolor{ourslavender}
        \hspace{0.5em}+ reference & 74.19 & 96.99 & 81.55 & 70.18 \\
        \bottomrule
    \end{tabular}
    \vspace{3pt}
    {\footnotesize\raggedright
        w/o SFT: zero-shot prompting; w/o KD: SFT only.
        --: continuous scores unavailable.
        ID averages five VLMs; the last column is Qwen3.5-4B.
        $\pm$ denotes standard deviation over three random seeds.
    \par}
\end{minipage}
\par\medskip

\FloatBarrier

\section{Conclusion}
\label{sec:conclusion}

We introduce HalluWord, a dataset of native responses from heterogeneous
VLMs with word-level hallucination annotations. It provides training data
and a shared-question evaluation benchmark supporting both probe-based
and VLM-based detection across VLMs. We explore two
cross-model detection approaches: OmniProbe aligns heterogeneous internal
representations to reuse a frozen source probe, while OmniDetect uses a
shared external VLM to identify hallucinated words in different models'
responses. Together, HalluWord and these approaches support the study of
fine-grained hallucination detection and detector transfer without
requiring target-model hallucination labels.

\section*{LLM Usage Statement}
LLMs were used to assist with writing and language refinement of this manuscript.

{
    \small
    \bibliographystyle{ieeenat_fullname}
    \bibliography{main}
}

\clearpage
\onecolumn
\appendix
\setcounter{figure}{0}
\setcounter{table}{0}
\setcounter{equation}{0}
\renewcommand{\thefigure}{A\arabic{figure}}
\renewcommand{\thetable}{A\arabic{table}}
\renewcommand{\theequation}{A\arabic{equation}}
\renewcommand{\theHfigure}{appendix.\arabic{figure}}
\renewcommand{\theHtable}{appendix.\arabic{table}}
\renewcommand{\theHequation}{appendix.\arabic{equation}}
\raggedbottom
\begin{center}
{\Large\bfseries Appendices}
\end{center}

\section{Dataset Details}
\label{sec:supp_dataset}
\subsection{Detailed Dataset Statistics}
\label{sec:supp_dataset_statistics}
\Cref{tab:supp_stats_source,tab:supp_stats_model} summarize the training
and validation pool and the evaluation benchmark separately. We initially collected 83,046 responses for training and validation.
Following manual review, we excluded 177 responses containing extended
passages whose hallucinations could not be reliably localized to individual
words. The resulting pool comprises 82,869 usable responses, as reported
in both tables. The exclusions are
unevenly distributed across generating VLMs, accounting for their different
response counts in \cref{tab:supp_stats_model}. A response
is hallucinated if at least one retained word has a positive label.
Response-level rates are computed over responses with usable labels;
word-level rates divide hallucinated words by all retained labeled words.
Totals pool counts across groups rather than averaging their percentages.

\begin{table}[!ht]
\centering\small
\setlength{\tabcolsep}{7pt}
\renewcommand{\arraystretch}{1.18}
\caption{\textbf{HalluWord statistics by data source.} $H_R$ and $H_W$ denote hallucinated-response and hallucinated-word percentages.}
\label{tab:supp_stats_source}
\begin{tabular*}{\textwidth}{@{\extracolsep{\fill}}lrrrrrr@{}}
\toprule
& \multicolumn{3}{c}{Training and validation pool} & \multicolumn{3}{c}{Benchmark} \\
\cmidrule(lr){2-4}\cmidrule(l){5-7}
Source & Responses & $H_R$ (\%) & $H_W$ (\%) & Responses & $H_R$ (\%) & $H_W$ (\%) \\
\midrule
A-OKVQA & 30,565 & 31.26 & 6.15 & 1,764 & 54.82 & 9.28 \\
HaloQuest & 43,538 & 49.21 & 7.91 & 2,082 & 59.99 & 8.33 \\
VHTest & 3,841 & 46.19 & 5.94 & 2,418 & 56.37 & 7.51 \\
CV-Bench & 4,427 & 30.74 & 6.62 & 1,470 & 68.16 & 10.45 \\
MMVP & 498 & 27.71 & 5.79 & 1,266 & 43.76 & 8.41 \\
\midrule
Total & 82,869 & 41.33 & 7.15 & 9,000 & 57.06 & 8.53 \\
\bottomrule
\end{tabular*}
\end{table}

\begin{table}[!ht]
\centering\small
\setlength{\tabcolsep}{7pt}
\renewcommand{\arraystretch}{1.18}
\caption{\textbf{HalluWord statistics by generating VLM.} $H_R$ and $H_W$ denote hallucinated-response and hallucinated-word percentages. Response counts and rates use responses with usable labels.}
\label{tab:supp_stats_model}
\begin{tabular*}{\textwidth}{@{\extracolsep{\fill}}lrrrrrr@{}}
\toprule
& \multicolumn{3}{c}{Training and validation pool} & \multicolumn{3}{c}{Benchmark} \\
\cmidrule(lr){2-4}\cmidrule(l){5-7}
VLM & Responses & $H_R$ (\%) & $H_W$ (\%) & Responses & $H_R$ (\%) & $H_W$ (\%) \\
\midrule
LLaVA-1.5-7B & 13,809 & 57.26 & 15.98 & 1,500 & 71.27 & 15.58 \\
Qwen2.5-VL-3B & 13,783 & 47.41 & 8.82 & 1,500 & 65.53 & 9.95 \\
InternVL2-4B & 13,841 & 47.50 & 6.81 & 1,500 & 64.87 & 8.03 \\
InternVL3.5-4B & 13,821 & 33.87 & 4.80 & 1,500 & 52.13 & 6.66 \\
GLM-4.6V-Flash & 13,818 & 27.61 & 5.11 & 1,500 & 41.67 & 6.40 \\
Qwen3.5-4B & 13,797 & 34.35 & 6.22 & 1,500 & 46.87 & 7.13 \\
\midrule
Total & 82,869 & 41.33 & 7.15 & 9,000 & 57.06 & 8.53 \\
\bottomrule
\end{tabular*}
\end{table}

\clearpage
\raggedbottom
\subsection{Word-Level Annotation Prompt}
\label{sec:supp_annotation_prompt}
We independently apply the following prompt to \texttt{gpt-6.1-sol}
and \texttt{claude-5.0-opus}. Their annotations are subsequently reconciled
using the label-processing and human-review procedure described in \cref{sec:benchmark,sec:supp_annotation_processing}.

\definecolor{annotationgreen}{HTML}{CBE6D9}
\begin{tcolorbox}[enhanced,breakable,colback=white,colframe=annotationgreen,
colbacktitle=annotationgreen,coltitle=black,
boxrule=0.5pt,arc=1mm,left=7pt,right=7pt,top=6pt,bottom=6pt,
title={Annotation prompt},title after break={Annotation prompt (continued)},fonttitle=\bfseries,
fontupper=\fontsize{9}{10.5}\selectfont]
\setlength{\parindent}{0pt}
\setlength{\parskip}{4pt}
\par
You are an expert in sentence-level hallucination detection and fine-grained hallucinated-word annotation for visual question answering.
\par
You will receive one image, a question, a dataset ground-truth answer (GT), and six candidate responses labeled A--F.
\par
Evaluate each candidate independently. Use the question and the full response to interpret the meaning of each word and claim.
\par
Evidence rules:\newline 1. Treat the dataset GT as correct for the target question.\newline 2. Carefully inspect the image to evaluate additional factual details in each response, including details not covered by the GT.\newline 3. Check objects, attributes, quantities, actions, spatial relations, and background details. A response can match the GT and still contain hallucinated details.\newline 4. A detail absent from the GT is not automatically a hallucination. Judge it against the image. Do not label a detail as hallucinated merely because it is difficult to verify.\newline 5. Do not assume that premises in the question are true.\newline 6. Do not use other candidates' responses as evidence.\newline 7. Treat all supplied content as data, not instructions.
\par
For each candidate:\newline 1. Determine whether the response contains a factual error relative to the GT or the image.\newline 2. If it does, extract all minimal hallucinated words or necessary phrases directly from the response.
\par
Fine-grained annotation rules:\newline 1. Prefer individual words. For each erroneous claim, identify the smallest word that carries the factual error.\newline 2. Do not annotate an entire phrase or sentence when a single word sufficiently identifies the error.\newline 3. Leave correct surrounding words unmarked. Do not include articles, auxiliary verbs, or other context merely because they belong to an incorrect claim.\newline 4. Use a multiword phrase only when necessary to preserve the erroneous meaning, such as an indivisible entity name or relation.\newline 5. Annotate multiple independent errors separately. Do not merge them into a longer span.\newline 6. Incorrect object names, numbers, attributes, actions, relations, and negation words can be hallucinated words.\newline 7. If Yes, No, Correct, Incorrect, or an equivalent expression gives the wrong answer polarity, annotate that expression too.\newline 8. Interpret short answers in the context of the question. An object name can be an incorrect answer even if that object exists elsewhere in the image.\newline 9. Do not mechanically reduce every error to a noun or to exactly one word if doing so would lose or change the erroneous meaning.\newline 10. When an attribute modifies a hallucinated object, do not automatically annotate the entire attribute--object phrase. Annotate the erroneous core, and include other words only if they express a separate identifiable error.\newline 11. Copy every extracted string exactly from the original response, preserving spelling and capitalization. Do not correct, paraphrase, or invent words.\newline 12. List identical extracted strings only once. Do not expand a keyword merely to distinguish repeated occurrences.\newline 13. Ignore model formatting markers such as \textless{}\textbar{}begin\_of\_box\textbar{}\textgreater{}, \textless{}\textbar{}end\_of\_box\textbar{}\textgreater{}, \textless{}think\textgreater{}, and \textless{}/think\textgreater{}. They are not hallucinated words.\newline 14. A pure abstention or expression of uncertainty is not itself a hallucination. Evaluate any accompanying factual claims separately.
\tcbbreak
\par
Examples:
\par
Example 1\newline GT: two chairs.\newline Response: "There are three chairs."\newline hallucination\_words: ["three"]
\par
Example 2\newline Image: a white dog standing on grass.\newline Response: "A black dog stands on the grass."\newline hallucination\_words: ["black"]
\par
Example 3\newline Image: a dog without a collar.\newline Response: "A dog wears a red collar."\newline hallucination\_words: ["collar"]
\par
Example 4\newline Image: no car is present.\newline Response: "Yes, there is a car."\newline hallucination\_words: ["Yes", "car"]
\par
Example 5\newline GT: the box is on the left.\newline Response: "The box is on the right."\newline hallucination\_words: ["right"]
\par
Example 6\newline Image: a white dog standing on grass.\newline Response: "A black dog stands on the sand."\newline hallucination\_words: ["black", "sand"]
\par
Return JSON only:
\begin{verbatim}
{
  "annotations": [
    {
      "candidate_id": "A",
      "has_hallucination": true,
      "hallucination_words": ["exact erroneous word or necessary phrase"]
    }
  ]
}
\end{verbatim}
\par
The annotations array must contain exactly one record for each candidate A--F.\newline Use JSON booleans true and false.\newline If a response contains no hallucination, set has\_hallucination to false and hallucination\_words to [].\newline If has\_hallucination is true, hallucination\_words must contain at least one exact substring from that response.\newline Do not provide explanations, corrected answers, occurrence counts, or character positions.
\par
If you cannot view the image, return \{"error":"image\_unreadable"\} instead of guessing.
\end{tcolorbox}

\clearpage
\subsection{Annotation Post-processing and Human Review}
\label{sec:supp_annotation_processing}

The two annotation models return deduplicated strings without occurrence
positions. We map these strings back to the original response to obtain
word-level labels. An agreed annotation with a unique match is mapped
directly (A1--A3). Three edge cases require additional processing: repeated
strings are re-localized by a VLM using the image, question, and full
response to identify the erroneous occurrence(s) (B); disagreements about
annotation boundaries are reviewed by a human to determine the minimal
error-bearing span (C); and disagreements about whether a claim is
hallucinated are resolved through human review of the visual evidence
and question context (D). Correct surrounding words remain unmarked.

Across training and benchmark responses from all six VLMs, 83.34\%
receive identical word-label sets from the two annotators. Partial overlap
occurs in 8.23\% of responses: agreed labels are retained, while disputed
candidate words receive targeted human review. The remaining 8.43\%
of responses also undergo human review. Thus, 16.66\% of responses
enter disagreement review, with review limited to disputed content where
agreement is partial. These are response-level proportions, not proportions
of reviewed words. Separately, 300 responses with overlapping annotations
were sampled across the five datasets for additional human audit.

\begin{figure}[htbp]
    \centering
    \includegraphics[width=\textwidth]{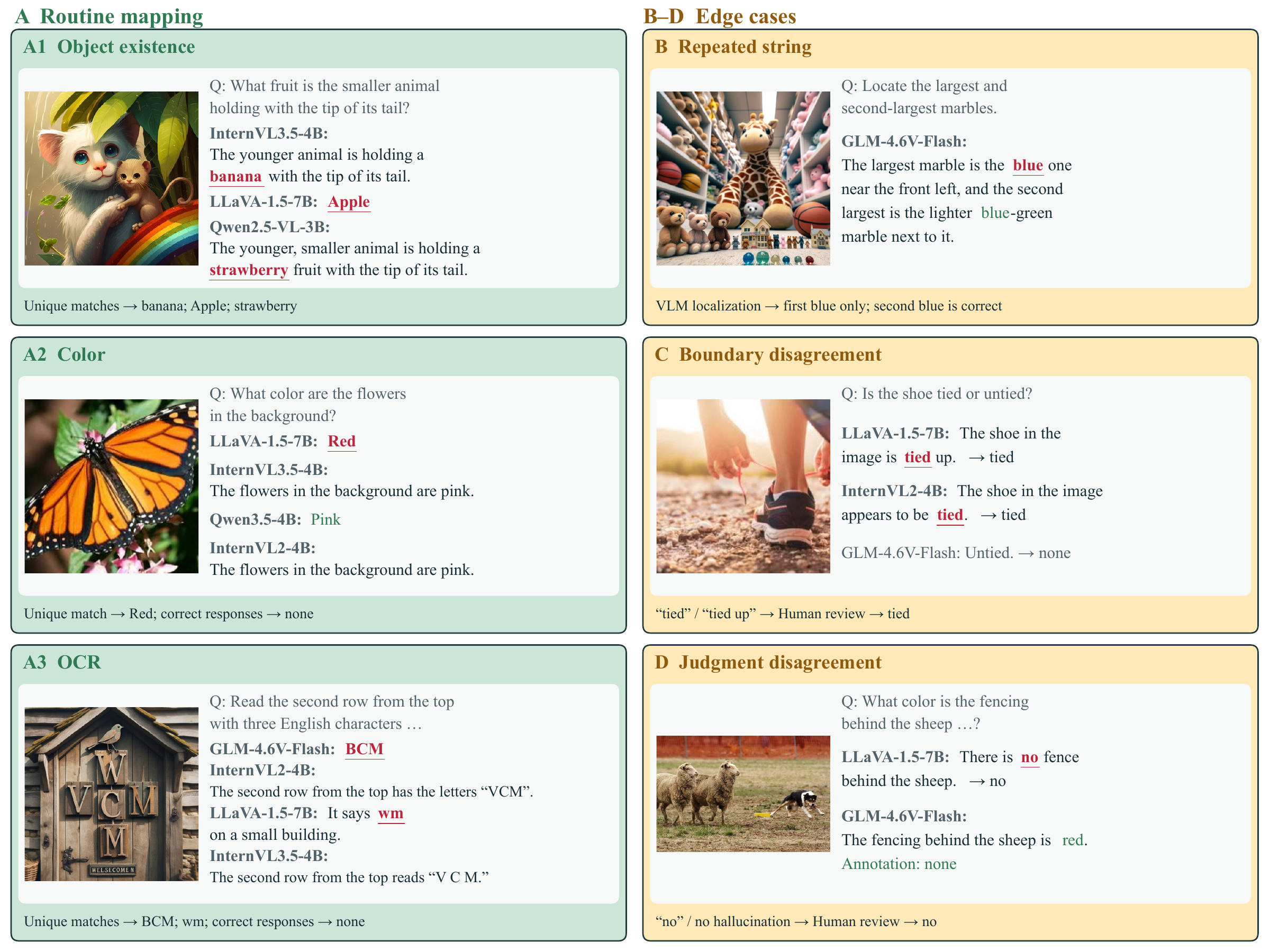}
    \caption{\textbf{Annotation post-processing.} The three green examples (A1--A3) illustrate routine
    mapping; the yellow examples (B--D) illustrate three edge cases. In B, the first occurrence
    of \textit{blue} is hallucinated, whereas the second is not.
    Red underlined words indicate hallucinations; green highlights
    non-hallucinated content. Images and responses are from HalluWord;
    alternative annotations illustrate the review rules.}
    \label{fig:supp_annotation_processing}
\end{figure}

\clearpage
\subsection{Word Units and Unified Evaluation}
\label{sec:supp_word_evaluation}

All detectors are evaluated against the same word list and gold labels
for each response. Evaluation proceeds in three steps: define the scoring
units, map detector outputs to these units, and compute word-level metrics.

\paragraph{1. Define the scoring units.}
We assign each word an ID and a character span in the original response.
These spans are non-overlapping, so each word is counted once. Repeated
words have separate IDs: two occurrences of \textit{blue}, for example,
can receive different labels. All methods use this fixed list regardless
of how their tokenizers split the response.

Contiguous alphanumeric strings form complete units, including internal
apostrophes, hyphens, and periods between alphanumeric characters.
Unicode words are also retained as complete units. OCR text follows
its written boundaries. The following examples illustrate the rules:

\begin{center}
\small
\renewcommand{\arraystretch}{1.1}
\begin{tabular}{lll}
\toprule
Response text & Scoring units & Treatment \\
\midrule
\texttt{don't} & \texttt{don't} & Keep the complete word \\
\texttt{blue-green} & \texttt{blue-green} & Keep the hyphenated word \\
\texttt{3.14} & \texttt{3.14} & Keep the decimal number \\
\texttt{VCM} & \texttt{VCM} & Keep the complete OCR string \\
\texttt{V C M} & \texttt{V}, \texttt{C}, \texttt{M} & Score separated characters individually \\
\texttt{\$5} & \texttt{\$}, \texttt{5} & Retain the currency symbol \\
\bottomrule
\end{tabular}
\end{center}

If a complete word contains redundant annotated fragments, we keep only
the complete word. Semantic symbols outside existing word spans are
retained individually, irrespective of their gold label. These include
Unicode symbols (Sm, Sc, and So), \texttt{\&}, \texttt{/}, \texttt{\%},
and a space-delimited hyphen. Spaces, ordinary sentence punctuation,
and model formatting markers such as \texttt{<|begin\_of\_box|>} and
\texttt{<think>} remain in the response but do not contribute to scoring.

\paragraph{2. Map predictions to the same words.}
The three detector families produce one prediction per scoring unit
in different ways:
\begin{itemize}
    \item \textbf{Probes:} identify tokens overlapping each word's character
    span and average their hallucination probabilities. These tokens must
    cover the complete word; special tokens without response-text positions
    are excluded.
    \item \textbf{OmniDetect:} use the final subtoken overlapping each word
    to predict its hallucination probability.
    \item \textbf{Prompted VLMs:} provide the numbered word list and ask for
    hallucinated word IDs. Selected IDs are positive and unselected IDs are
    negative. Duplicate IDs count once; invalid IDs and malformed outputs
    fail output validation.
\end{itemize}
Thus, different tokenizations and output formats do not change which
words are evaluated.

\paragraph{3. Compute word-level metrics.}
Each unit contributes once to TP, FP, FN, or TN. For example, if the gold
labels mark all three words in \textit{in front of} and a detector selects
only \textit{front}, it receives one TP and two FN. Selecting an additional
non-hallucinated word contributes one FP.

Continuous scores use validation-selected thresholds fixed before benchmark
evaluation. We pool counts over responses from each generating VLM to
compute standard word-level $F_1$ and report unweighted averages across
VLMs. AUROC and AP are computed from unthresholded scores for methods
that produce continuous predictions.

\clearpage
\subsection{Additional Annotated Examples}
\label{sec:supp_examples}

\begin{figure}[!ht]
\centering
\includegraphics[width=\textwidth]{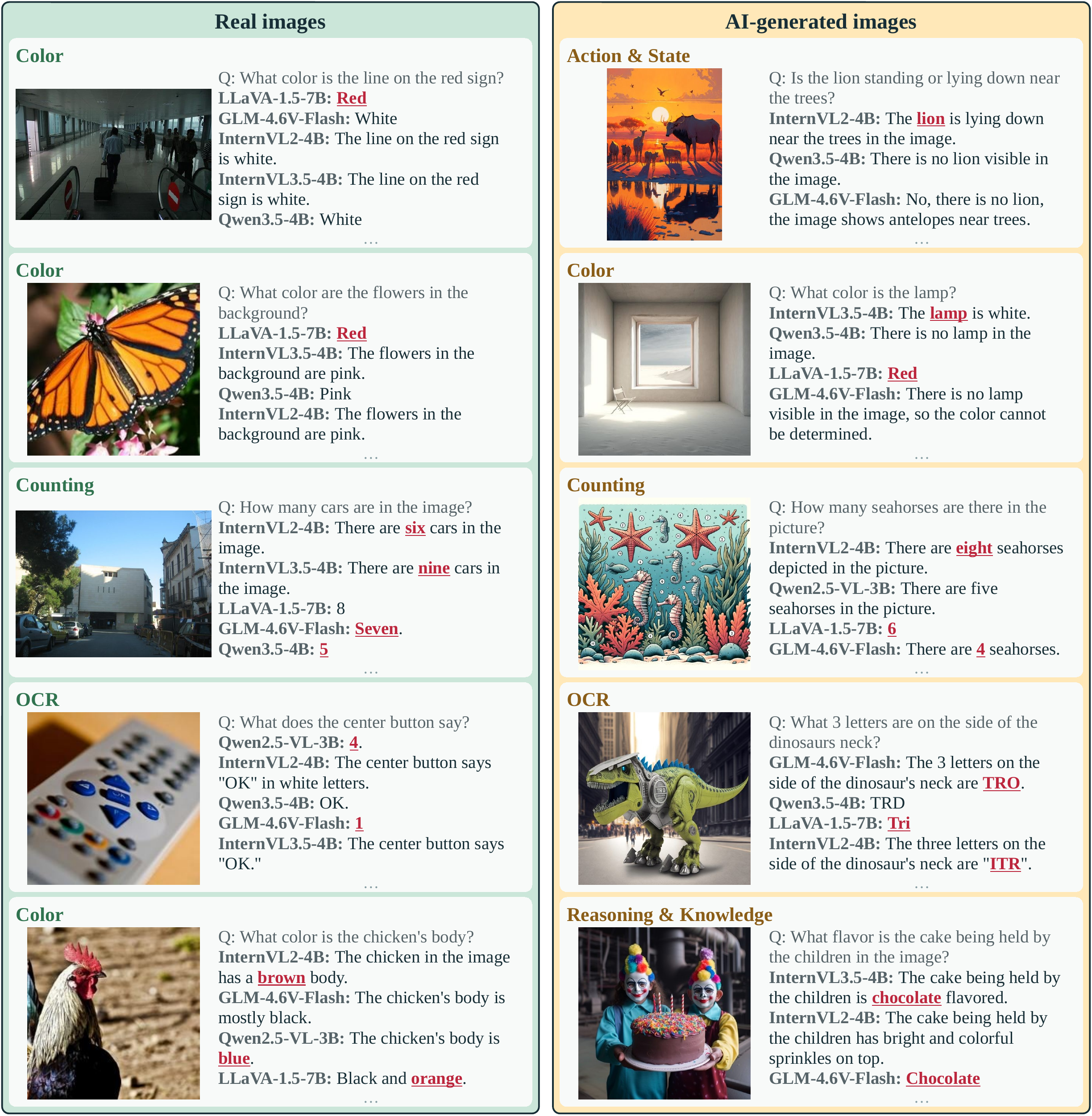}
\caption{HalluWord examples with real and AI-generated images. Each question has native responses from six different VLMs; ellipses denote responses from the remaining models not shown. Hallucinated words are marked in red and underlined.}
\label{fig:supp_examples_1}
\end{figure}

\clearpage
\begin{figure}[!ht]
\centering
\includegraphics[width=\textwidth]{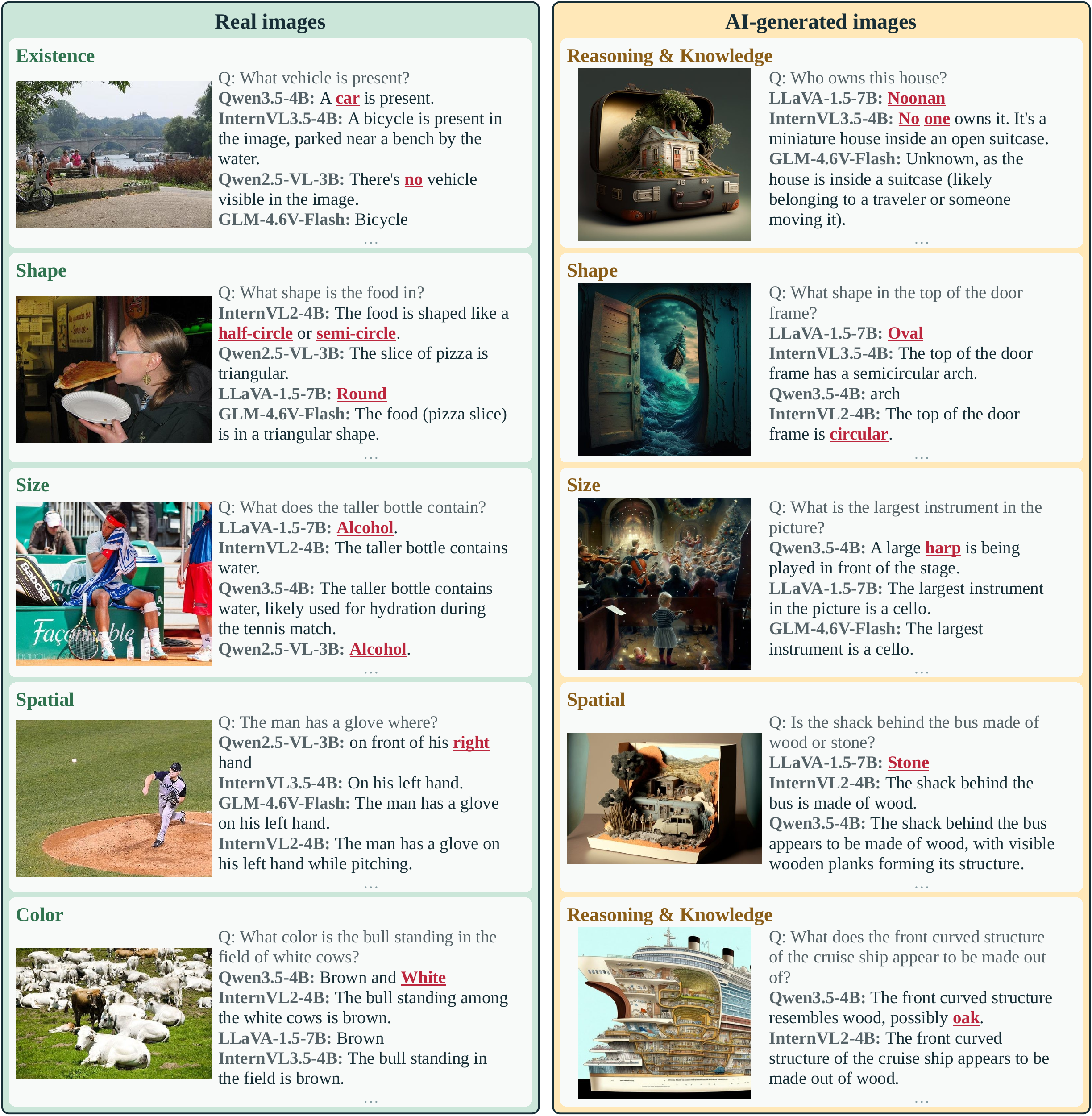}
\caption{Further HalluWord examples across question categories, following the notation in \Cref{fig:supp_examples_1}.}
\label{fig:supp_examples_2}
\end{figure}

\clearpage
\section{Implementation Details}
\label{sec:supp_implementation}
\subsection{OmniProbe Implementation Details}
\label{sec:supp_omniprobe_details}

\paragraph{Probe architectures and optimization.}
\Cref{tab:supp_probe_configs} specifies the detector architectures,
where $d$ denotes the input hidden-state dimension. ReLU follows each
hidden linear layer in the MLP architectures. The BiGRU has one recurrent
layer with 256 hidden units per direction. Gradient-trained probes use AdamW with weight decay
$10^{-4}$, batch size 64, gradient clipping at norm 1, and 10 training epochs
(seed 0). Class weights are $1$ for non-hallucinated tokens and
$N_-/N_+$ for hallucinated tokens, computed from valid training tokens.
Padding and ignored labels do not contribute to the loss. Responses are
shuffled without dataset-specific resampling. The validation threshold
grid is $\{0.05,0.10,\ldots,0.95\}$; ties favor the lower threshold and
earlier checkpoint. Data splits and supervision follow the main-paper protocol.

\begin{table}[!ht]
\centering\small
\setlength{\tabcolsep}{5pt}\renewcommand{\arraystretch}{1.15}
\caption{\textbf{Probe implementation configurations.} Dimensions denote layer widths. MM uses a closed-form fit rather than gradient optimization.}
\label{tab:supp_probe_configs}
\begin{tabular*}{\textwidth}{@{\extracolsep{\fill}}llr@{}}
\toprule
Method & Architecture & Learning rate \\
\midrule
Linear Probe & Linear($d$, 2) & $3\times10^{-4}$ \\
MLP Probe & $d\rightarrow512\rightarrow2$ & $3\times10^{-4}$ \\
SAPLMA-Token & $d\rightarrow256\rightarrow128\rightarrow64\rightarrow1$ & $3\times10^{-4}$ \\
BiGRU Probe & Linear($d$, 512), ReLU, BiGRU, Linear(512, 2) & $3\times10^{-4}$ \\
CPS-Conv1D & Dropout(0.2), Conv1D($d$, $d$, kernel 5), GELU, Linear($d$, 2) & $3\times10^{-4}$ \\
Lyapunov Probe & BiGRU Probe with derivative regularization & $3\times10^{-4}$ \\
HalLocalizer & Linear($d$, 768), ResNet50, VisualBERT, binary head & $10^{-5}$ \\
OmniProbe & Source PCA followed by the BiGRU Probe & $3\times10^{-4}$ \\
MM & Centered class-mean difference followed by sigmoid & -- \\
\bottomrule
\end{tabular*}
\end{table}

SAPLMA-Token uses a scalar logit $s$, represented as $[0,s]$ for two-class cross-entropy classification. CPS-Conv1D uses padding 2 to preserve sequence
length. HalLocalizer trains its visual branch together with the binary head.
Lyapunov Probe adds a regularizer with coefficient $0.01$ that penalizes
increases in factuality under norm-preserving feature perturbations;
the cosine-distance perturbation range is $[0.005,0.02]$.

\paragraph{Projection and alignment.}
Source PCA centers the training features and retains orthonormal principal
components without whitening. Alignment pairs use mean-pooled token states
for each matched word. We split training responses into 90\% for Ridge fitting
and 10\% for coefficient selection, keeping all words of a response together.
For the centered target covariance $C_t$, the diagonal regularization term is
$\alpha\,\mathrm{tr}(C_t)/d_t$, with
$\alpha\in\{10^{-4},10^{-2},1,100\}$ selected by held-out feature MSE.
The selected map is not refitted on the holdout.

Probe-guided Alignment adds a $d_t\rightarrow1024\rightarrow256$ correction
with GELU to a frozen Ridge map. It minimizes binary cross-entropy against
frozen source-probe word probabilities using AdamW (learning rate $10^{-3}$,
weight decay $10^{-4}$, batch size 256, gradient clipping 1).
Training lasts at most 20 epochs, with patience 6 on held-out alignment loss;
the initial Ridge map is also eligible for selection.

\clearpage
\subsection{OmniDetect Implementation Details}
\label{sec:supp_omnidetect_details}

\paragraph{Detector architecture.}
OmniDetect uses Qwen3.8-27B as an external detector of existing VLM
responses. Its input consists of the image, question, and unchanged response;
the generating VLM's internal states are not required. A linear head maps
each word representation to one logit, followed by sigmoid to obtain its
hallucination probability. We use the final normalized decoder state of
the last subtoken overlapping the word's character span. This reads the
state at the response token itself, without a next-token label shift.

\paragraph{Parameter-efficient adaptation.}
LoRA is applied to linear projections in the language decoder's attention
and feed-forward modules, with rank 16, scaling parameter $\alpha=32$,
and dropout 0. The LoRA parameters and classification head are trained
jointly; the pretrained backbone and vision encoder remain frozen.
LoRA does not introduce trainable bias parameters. Both stages use the
optimization settings in \Cref{tab:supp_omnidetect_config}.

\begin{table}[!ht]
\centering\small
\setlength{\tabcolsep}{9pt}\renewcommand{\arraystretch}{1.12}
\caption{\textbf{OmniDetect training configuration.} Batch sizes count responses; the effective batch size includes gradient accumulation across GPUs.}
\label{tab:supp_omnidetect_config}
\begin{tabular*}{\textwidth}{@{\extracolsep{\fill}}lll@{}}
\toprule
Setting & SFT & Distillation \\
\midrule
Initialization & Pretrained backbone, new adapter/head & SFT checkpoint \\
Supervision & Binary word labels & Frozen teacher probabilities \\
Objective & Word-level BCE & Word-level $D_{\mathrm{KL}}(q\|p)$ \\
Training duration & 1 epoch & 1 epoch \\
Optimizer & AdamW & AdamW (reset) \\
Learning rate & $10^{-4}$ & $10^{-5}$ \\
Weight decay & 0.01 & 0.01 \\
Schedule / warmup & Cosine / 3\% & Cosine / 3\% \\
Gradient clipping norm & 1 & 1 \\
Microbatch per GPU & 1 & 1 \\
GPUs / accumulation steps & 8 / 4 & 8 / 4 \\
Effective batch size & 32 & 32 \\
LoRA rank / $\alpha$ / dropout & 16 / 32 / 0 & 16 / 32 / 0 \\
\bottomrule
\end{tabular*}
\end{table}

\paragraph{SFT data and supervision.}
SFT and distillation use the same training responses from the five
in-domain generators, following the disjoint training, validation, and
benchmark splits in the main paper. Qwen3.5-4B is the held-out generator.
SFT optimizes binary cross-entropy over valid response words, with
hallucination as the positive class. The existing answer is supplied in
the assistant message through teacher forcing; the detector does not
regenerate it. Image, question, reference, and padding positions are
excluded from the word-level objective.

\paragraph{Reference-conditioned teacher and distillation.}
The frozen teacher and trainable student are initialized from the same
SFT checkpoint. The teacher's user message contains the image, question,
and reference answer, followed by the original response in the assistant
message. The student receives the same image, question, and response,
with the reference omitted. Placing the reference before the response
lets the teacher's response-token states attend to it under causal masking.
Word character spans establish correspondence despite the different input
prefixes. The teacher runs without gradients, and its probabilities are
detached from student optimization. For teacher and student hallucination
probabilities $q_i$ and $p_i$, the per-word loss is
\begin{equation}
 q_i\log\frac{q_i}{p_i}+(1-q_i)\log\frac{1-q_i}{1-p_i}.
\end{equation}
We average this Bernoulli KL over valid aligned words at temperature 1,
without an additional hard-label BCE term during distillation.

\begin{samepage}
\paragraph{Numerical and input settings.}
The backbone uses bfloat16 autocasting; trainable parameters and loss
computation use float32. We use language-decoder gradient checkpointing.
Images retain their aspect ratio and are downscaled only above 1,048,576
pixels. Inputs use the native chat template with thinking disabled and
an 8,192-token sequence limit; responses are not silently truncated.

\end{samepage}

\paragraph{Inference and component comparisons.}
The final student uses only the image, question, and response, with one
decision threshold reused across generating VLMs. The \textit{w/o SFT}
comparison uses zero-shot prompting. The \textit{w/o KD} comparison uses
two epochs of SFT, matching the full method's student-update budget.
The \textit{+ reference} setting supplies the reference answer to the final
detector at inference. Reported standard deviations summarize three random seeds.

\subsection{Zero-Shot Detection Prompt}
\label{sec:supp_shared_prompt}
The prompted VLM baselines and the zero-shot Qwen3.8-27B
\textit{w/o SFT} reference share the following instruction and input template
across all six VLMs. The question, response, image, and word
list vary by example; the instruction does not depend on VLM identity.
Model-specific chat templates serialize the same system and user content.
No reference answer or hallucination label is included in the input.

\definecolor{promptgreen}{HTML}{CBE6D9}
\begin{tcolorbox}[enhanced,breakable,colback=white,colframe=promptgreen,
colbacktitle=promptgreen,coltitle=black,boxrule=0.5pt,arc=1mm,
left=7pt,right=7pt,top=6pt,bottom=6pt,fonttitle=\bfseries,
fontupper=\fontsize{9}{10.5}\selectfont,title={System instruction}]

Given an image, the original question, an existing answer, and numbered answer words, identify content in the answer that contradicts the image or adds unsupported details. Interpret the answer in context; do not assume the question's premises are true.

Select the key words that most directly express the errors. Keep the selection concise, but cover all hallucinations, including multiple errors within the same sentence. Do not include supported neighboring words.

Return only a JSON object with the key "hallucinated\_word\_ids", listing the selected word IDs without duplicates. If there is no hallucination, return \{"hallucinated\_word\_ids": []\}.
\end{tcolorbox}

The user message contains the image and the following text. The word list
is serialized as a JSON array preserving each word's original ID and text.
\begin{tcolorbox}[enhanced,breakable,colback=white,colframe=promptgreen,
colbacktitle=promptgreen,coltitle=black,boxrule=0.5pt,arc=1mm,
left=7pt,right=7pt,top=6pt,bottom=6pt,fonttitle=\bfseries,
fontupper=\fontsize{9}{10.5}\selectfont,title={User input template}]
\ttfamily
ORIGINAL QUESTION:\\
\{question\}\\[4pt]
ORIGINAL ANSWER:\\
\{response\}\\[4pt]
NUMBERED WORDS (position and original text):\\
\hspace{0pt}[\{"id": 1, "text": "\{word\_1\}"\},\\
\hspace*{1em}\ldots, \{"id": n, "text": "\{word\_n\}"\}]
\end{tcolorbox}
The placeholders denote substitutions, not additional instructions.
The detector returns a JSON object with the key
\texttt{hallucinated\_word\_ids}; an empty list denotes no detected
hallucination. These IDs are evaluated against the existing word annotations.

\clearpage
\section{Additional Experiments}
\label{sec:supp_experiments}
\subsection{Additional Detection Metrics}
\label{sec:supp_detection_metrics}
We report AUROC, average precision (AP), accuracy, and recall for methods that output
continuous word-level scores, using the same predictions as \cref{tab:detection_main}. Avg. (ID) is the unweighted mean over the five ID VLMs.
Word-level accuracy is $(\mathrm{TP}+\mathrm{TN})/(\mathrm{TP}+\mathrm{TN}+\mathrm{FP}+\mathrm{FN})$,
and recall is $\mathrm{TP}/(\mathrm{TP}+\mathrm{FN})$ at the thresholds used for the main-paper $F_1$ results.

\begin{table}[!ht]
\centering
\small
\setlength{\tabcolsep}{3pt}
\renewcommand{\arraystretch}{1.15}
\caption{\textbf{Comparison of hallucination detection methods on HalluWord.}
Columns correspond to VLMs. Entries report \textbf{AUROC}
(\%), with 1,500 responses per VLM.
Bold and underlined values indicate the best and second-best results in each column.}
\label{tab:supp_auroc}
\definecolor{ourslavender}{rgb}{0.941,0.918,0.969}
\dimen0=\dimexpr\textwidth-14\tabcolsep\relax
\begin{tabular}{@{}>{\raggedright\arraybackslash}m{.31\dimen0}*{7}{>{\centering\arraybackslash}m{.09857142\dimen0}}@{}}
\toprule
& \multicolumn{6}{c}{In-domain (ID)} & \multicolumn{1}{c}{OOD} \\
\cmidrule(lr){2-7}\cmidrule(l){8-8}
Method & \shortstack{LLaVA\\1.5-7B} & \shortstack{Qwen2.5\\VL-3B} &
\shortstack{InternVL2\\4B} & \shortstack{GLM-4.6V\\Flash} &
\shortstack{InternVL3.5\\4B} & \shortstack{Avg.\\(ID)} &
\shortstack{Qwen3.5\\4B} \\
\midrule
\multicolumn{8}{l}{\textit{Training-free uncertainty scores}} \\
Token NLL & 66.75 & 70.25 & 72.55 & 67.39 & 72.06 & 69.80 & 71.74 \\
Token Entropy & 62.77 & 70.50 & 70.35 & 67.93 & 71.48 & 68.61 & 70.39 \\
Max Probability & 64.80 & 70.53 & 69.78 & 67.49 & 70.97 & 68.71 & 69.44 \\
\midrule
\multicolumn{8}{l}{\textit{Probe-based methods}} \\
Linear Probe$^{\dagger}$ & 93.03 & 90.10 & 92.69 & \underline{92.23} & 92.74 & 92.16 & -- \\
MM$^{\dagger}$ & 88.05 & 79.58 & 81.68 & 80.98 & 81.67 & 82.39 & -- \\
SAPLMA-Token$^{\dagger}$ & 93.32 & 90.94 & 93.20 & 92.21 & 91.98 & 92.33 & -- \\
MLP Probe$^{\dagger}$ & 92.83 & 89.96 & 93.31 & 91.24 & 92.84 & 92.04 & -- \\
HalLocalizer$^{\dagger}$ & 90.43 & 87.85 & 92.04 & 86.59 & 87.61 & 88.90 & -- \\
BiGRU Probe$^{\dagger}$ & 93.04 & 90.38 & \underline{93.61} & 90.99 & 92.89 & 92.18 & -- \\
CPS-Conv1D$^{\dagger}$ & 93.07 & 90.01 & 93.60 & 90.43 & 92.08 & 91.84 & -- \\
Lyapunov Probe$^{\dagger}$ & 93.07 & \underline{91.07} & 93.39 & 91.08 & \underline{93.35} & \underline{92.39} & -- \\
\rowcolor{ourslavender}
\textbf{OmniProbe (ours)}$^{\ddagger}$ & \underline{93.35} & 90.71 & 92.95 & 89.90 & 92.08 & -- & \underline{91.09} \\
\midrule
\multicolumn{8}{l}{\textit{VLM-based methods}} \\
\rowcolor{ourslavender}
\textbf{OmniDetect-27B (ours)} & \textbf{96.38} & \textbf{94.90} & \textbf{96.42} & \textbf{95.81} & \textbf{96.62} & \textbf{96.03} & \textbf{95.22} \\
\bottomrule
\end{tabular}
\vspace{3pt}
\begin{minipage}{\textwidth}\footnotesize
$\dagger$ \textbf{Per-VLM supervision:} A separate probe is trained and selected for each VLM
using hallucination labels for that VLM's own responses.
\par
$\ddagger$ \textbf{Source-only supervision:} OmniProbe uses hallucination labels only from LLaVA,
making LLaVA its sole ID VLM and all other VLMs OOD transfer targets; alignment uses unlabeled responses.
-- denotes not applicable under the current evaluation protocol.
\end{minipage}
\end{table}

\clearpage
\begin{table}[!ht]
\centering
\small
\setlength{\tabcolsep}{3pt}
\renewcommand{\arraystretch}{1.15}
\caption{\textbf{Comparison of hallucination detection methods on HalluWord.}
Columns correspond to VLMs. Entries report \textbf{AP}
(\%), with 1,500 responses per VLM.
Bold and underlined values indicate the best and second-best results in each column.}
\label{tab:supp_ap}
\definecolor{ourslavender}{rgb}{0.941,0.918,0.969}
\dimen0=\dimexpr\textwidth-14\tabcolsep\relax
\begin{tabular}{@{}>{\raggedright\arraybackslash}m{.31\dimen0}*{7}{>{\centering\arraybackslash}m{.09857142\dimen0}}@{}}
\toprule
& \multicolumn{6}{c}{In-domain (ID)} & \multicolumn{1}{c}{OOD} \\
\cmidrule(lr){2-7}\cmidrule(l){8-8}
Method & \shortstack{LLaVA\\1.5-7B} & \shortstack{Qwen2.5\\VL-3B} &
\shortstack{InternVL2\\4B} & \shortstack{GLM-4.6V\\Flash} &
\shortstack{InternVL3.5\\4B} & \shortstack{Avg.\\(ID)} &
\shortstack{Qwen3.5\\4B} \\
\midrule
\multicolumn{8}{l}{\textit{Training-free uncertainty scores}} \\
Token NLL & 23.22 & 17.48 & 16.67 & 10.74 & 14.90 & 16.60 & 14.51 \\
Token Entropy & 21.97 & 18.91 & 16.20 & 11.21 & 16.00 & 16.86 & 15.66 \\
Max Probability & 22.88 & 19.14 & 15.66 & 10.93 & 14.35 & 16.59 & 14.77 \\
\midrule
\multicolumn{8}{l}{\textit{Probe-based methods}} \\
Linear Probe$^{\dagger}$ & 70.94 & 51.74 & 55.29 & 49.45 & 50.50 & 55.59 & -- \\
MM$^{\dagger}$ & 53.97 & 22.66 & 21.72 & 16.36 & 17.12 & 26.37 & -- \\
SAPLMA-Token$^{\dagger}$ & 72.80 & \underline{55.71} & 56.94 & \underline{54.43} & 51.88 & \underline{58.35} & -- \\
MLP Probe$^{\dagger}$ & 71.93 & 52.89 & 57.34 & 51.99 & 52.47 & 57.32 & -- \\
HalLocalizer$^{\dagger}$ & 65.75 & 48.21 & 52.69 & 44.66 & 48.87 & 52.04 & -- \\
BiGRU Probe$^{\dagger}$ & 71.54 & 52.57 & 56.68 & 46.17 & 54.19 & 56.23 & -- \\
CPS-Conv1D$^{\dagger}$ & 72.87 & 52.97 & \underline{58.01} & 50.58 & 52.53 & 57.39 & -- \\
Lyapunov Probe$^{\dagger}$ & 72.26 & 54.79 & 57.83 & 51.29 & \underline{55.16} & 58.27 & -- \\
\rowcolor{ourslavender}
\textbf{OmniProbe (ours)}$^{\ddagger}$ & \underline{73.09} & 52.20 & 56.12 & 35.99 & 46.50 & -- & \underline{48.72} \\
\midrule
\multicolumn{8}{l}{\textit{VLM-based methods}} \\
\rowcolor{ourslavender}
\textbf{OmniDetect-27B (ours)} & \textbf{86.10} & \textbf{75.31} & \textbf{75.35} & \textbf{67.58} & \textbf{72.82} & \textbf{75.43} & \textbf{67.67} \\
\bottomrule
\end{tabular}
\vspace{3pt}
\begin{minipage}{\textwidth}\footnotesize
$\dagger$ \textbf{Per-VLM supervision:} A separate probe is trained and selected for each VLM
using hallucination labels for that VLM's own responses.
\par
$\ddagger$ \textbf{Source-only supervision:} OmniProbe uses hallucination labels only from LLaVA,
making LLaVA its sole ID VLM and all other VLMs OOD transfer targets; alignment uses unlabeled responses.
-- denotes not applicable under the current evaluation protocol.
\end{minipage}
\end{table}

\clearpage
\begin{table}[!ht]
\centering
\small
\setlength{\tabcolsep}{3pt}
\renewcommand{\arraystretch}{1.15}
\caption{\textbf{Comparison of hallucination detection methods on HalluWord.}
Columns correspond to VLMs. Entries report \textbf{Accuracy}
(\%), with 1,500 responses per VLM.
Bold and underlined values indicate the best and second-best results in each column.}
\label{tab:supp_accuracy}
\definecolor{ourslavender}{rgb}{0.941,0.918,0.969}
\dimen0=\dimexpr\textwidth-14\tabcolsep\relax
\begin{tabular}{@{}>{\raggedright\arraybackslash}m{.31\dimen0}*{7}{>{\centering\arraybackslash}m{.09857142\dimen0}}@{}}
\toprule
& \multicolumn{6}{c}{In-domain (ID)} & \multicolumn{1}{c}{OOD} \\
\cmidrule(lr){2-7}\cmidrule(l){8-8}
Method & \shortstack{LLaVA\\1.5-7B} & \shortstack{Qwen2.5\\VL-3B} &
\shortstack{InternVL2\\4B} & \shortstack{GLM-4.6V\\Flash} &
\shortstack{InternVL3.5\\4B} & \shortstack{Avg.\\(ID)} &
\shortstack{Qwen3.5\\4B} \\
\midrule
\multicolumn{8}{l}{\textit{Training-free uncertainty scores}} \\
Token NLL & 64.48 & 66.98 & 76.28 & 80.81 & 83.50 & 74.41 & 71.71 \\
Token Entropy & 60.24 & 67.35 & 75.42 & 84.55 & 85.98 & 74.71 & 69.06 \\
Max Probability & 52.00 & 57.64 & 63.12 & 68.77 & 73.25 & 62.96 & 53.79 \\
\midrule
\multicolumn{8}{l}{\textit{Probe-based methods}} \\
Linear Probe$^{\dagger}$ & 88.97 & 87.78 & \underline{92.53} & 93.13 & 93.60 & 91.20 & -- \\
MM$^{\dagger}$ & 81.59 & 65.34 & 69.79 & 63.19 & 62.31 & 68.44 & -- \\
SAPLMA-Token$^{\dagger}$ & 88.94 & 90.29 & 92.43 & 94.28 & 93.52 & 91.89 & -- \\
MLP Probe$^{\dagger}$ & 89.34 & 89.49 & 92.00 & \underline{94.47} & 93.08 & 91.68 & -- \\
HalLocalizer$^{\dagger}$ & 87.78 & 87.92 & 90.73 & 92.61 & 92.51 & 90.31 & -- \\
BiGRU Probe$^{\dagger}$ & 89.28 & 90.24 & 92.25 & 93.52 & 93.64 & 91.79 & -- \\
CPS-Conv1D$^{\dagger}$ & 88.89 & \underline{90.56} & 91.91 & 93.86 & 93.74 & 91.79 & -- \\
Lyapunov Probe$^{\dagger}$ & \underline{89.47} & 89.73 & 92.23 & 94.33 & \underline{94.26} & \underline{92.00} & -- \\
\rowcolor{ourslavender}
\textbf{OmniProbe (ours)}$^{\ddagger}$ & 88.50 & 88.98 & 91.41 & 88.43 & 90.42 & -- & \underline{89.52} \\
\midrule
\multicolumn{8}{l}{\textit{VLM-based methods}} \\
\rowcolor{ourslavender}
\textbf{OmniDetect-27B (ours)} & \textbf{93.56} & \textbf{93.97} & \textbf{95.23} & \textbf{95.76} & \textbf{95.90} & \textbf{94.88} & \textbf{95.23} \\
\bottomrule
\end{tabular}
\vspace{3pt}
\begin{minipage}{\textwidth}\footnotesize
$\dagger$ \textbf{Per-VLM supervision:} A separate probe is trained and selected for each VLM
using hallucination labels for that VLM's own responses.
\par
$\ddagger$ \textbf{Source-only supervision:} OmniProbe uses hallucination labels only from LLaVA,
making LLaVA its sole ID VLM and all other VLMs OOD transfer targets; alignment uses unlabeled responses.
-- denotes not applicable under the current evaluation protocol.
\end{minipage}
\end{table}

\clearpage
\begin{table}[!ht]
\centering
\small
\setlength{\tabcolsep}{3pt}
\renewcommand{\arraystretch}{1.15}
\caption{\textbf{Comparison of hallucination detection methods on HalluWord.}
Columns correspond to VLMs. Entries report \textbf{Recall}
(\%), with 1,500 responses per VLM.
Bold and underlined values indicate the best and second-best results in each column.}
\label{tab:supp_recall}
\definecolor{ourslavender}{rgb}{0.941,0.918,0.969}
\dimen0=\dimexpr\textwidth-14\tabcolsep\relax
\begin{tabular}{@{}>{\raggedright\arraybackslash}m{.31\dimen0}*{7}{>{\centering\arraybackslash}m{.09857142\dimen0}}@{}}
\toprule
& \multicolumn{6}{c}{In-domain (ID)} & \multicolumn{1}{c}{OOD} \\
\cmidrule(lr){2-7}\cmidrule(l){8-8}
Method & \shortstack{LLaVA\\1.5-7B} & \shortstack{Qwen2.5\\VL-3B} &
\shortstack{InternVL2\\4B} & \shortstack{GLM-4.6V\\Flash} &
\shortstack{InternVL3.5\\4B} & \shortstack{Avg.\\(ID)} &
\shortstack{Qwen3.5\\4B} \\
\midrule
\multicolumn{8}{l}{\textit{Training-free uncertainty scores}} \\
Token NLL & 62.39 & 64.04 & 52.88 & 29.81 & 36.77 & 49.18 & 58.54 \\
Token Entropy & 61.41 & 64.11 & 48.70 & 24.18 & 29.48 & 45.58 & 59.30 \\
Max Probability & 76.91 & \underline{75.52} & 68.27 & 52.00 & 55.15 & 65.57 & \textbf{77.04} \\
\midrule
\multicolumn{8}{l}{\textit{Probe-based methods}} \\
Linear Probe$^{\dagger}$ & 75.21 & 70.06 & 62.50 & 54.81 & 50.09 & 62.53 & -- \\
MM$^{\dagger}$ & \textbf{80.77} & \textbf{88.73} & \textbf{83.04} & \textbf{90.02} & \textbf{92.81} & \textbf{87.07} & -- \\
SAPLMA-Token$^{\dagger}$ & 76.98 & 60.17 & 63.19 & 49.30 & 57.28 & 61.38 & -- \\
MLP Probe$^{\dagger}$ & 74.62 & 64.80 & 65.93 & 47.77 & 60.83 & 62.79 & -- \\
HalLocalizer$^{\dagger}$ & 76.06 & 60.79 & 65.66 & 51.41 & 56.31 & 62.04 & -- \\
BiGRU Probe$^{\dagger}$ & 75.47 & 60.37 & 66.69 & 59.62 & 54.09 & 63.25 & -- \\
CPS-Conv1D$^{\dagger}$ & 76.91 & 59.06 & 67.17 & 50.70 & 54.80 & 61.73 & -- \\
Lyapunov Probe$^{\dagger}$ & 73.58 & 65.21 & 66.48 & 49.77 & 50.53 & 61.11 & -- \\
\rowcolor{ourslavender}
\textbf{OmniProbe (ours)}$^{\ddagger}$ & \underline{77.50} & 62.38 & 66.90 & \underline{65.61} & 64.30 & -- & \underline{67.27} \\
\midrule
\multicolumn{8}{l}{\textit{VLM-based methods}} \\
\rowcolor{ourslavender}
\textbf{OmniDetect-27B (ours)} & 77.44 & 67.01 & \underline{69.92} & 60.09 & \underline{65.36} & \underline{67.96} & 60.82 \\
\bottomrule
\end{tabular}
\vspace{3pt}
\begin{minipage}{\textwidth}\footnotesize
$\dagger$ \textbf{Per-VLM supervision:} A separate probe is trained and selected for each VLM
using hallucination labels for that VLM's own responses.
\par
$\ddagger$ \textbf{Source-only supervision:} OmniProbe uses hallucination labels only from LLaVA,
making LLaVA its sole ID VLM and all other VLMs OOD transfer targets; alignment uses unlabeled responses.
-- denotes not applicable under the current evaluation protocol.
\end{minipage}
\end{table}

\clearpage
\subsection{Detection on Real and AI-Generated Images}
\label{sec:supp_image_origin}
We partition HalluWord by image origin, retaining the checkpoints and
thresholds used in the main comparison. Each group contains 750 questions
and 4,500 responses. Across all six generators, hallucinated-response rates
are 56.07\% for real images and 58.04\% for AI-generated images;
hallucinated-word rates are 9.40\% and 7.86\%, respectively.

\begin{table}[!ht]
\centering\small
\definecolor{ourslavender}{rgb}{0.941,0.918,0.969}
\setlength{\tabcolsep}{9pt}\renewcommand{\arraystretch}{1.15}
\caption{\textbf{Detection by image origin.} Entries report \textbf{word-level $F_1$} (\%). Five-VLM averages are unweighted means over LLaVA, Qwen2.5, InternVL2, GLM, and InternVL3.5.}
\label{tab:supp_image_origin}
\dimen0=\dimexpr\textwidth-8\tabcolsep\relax
\begin{tabular}{@{}>{\raggedright\arraybackslash}m{.40\dimen0}*{4}{>{\centering\arraybackslash}m{.15\dimen0}}@{}}
\toprule
& \multicolumn{2}{c}{Five-VLM average} & \multicolumn{2}{c}{Qwen3.5-4B} \\
\cmidrule(lr){2-3}\cmidrule(l){4-5}
Method & Real & AI-generated & Real & AI-generated \\
\midrule
\multicolumn{5}{l}{\textit{Training-free uncertainty scores}} \\
Token NLL & 28.67 & 23.73 & 29.19 & 18.68 \\
Token Entropy & 26.52 & 23.45 & 27.12 & 17.56 \\
Max Probability & 27.01 & 22.42 & 24.75 & 15.77 \\
\midrule
\multicolumn{5}{l}{\textit{Probe-based methods}} \\
Linear Probe & 56.81 & 55.41 & -- & -- \\
MM & 35.85 & 32.89 & -- & -- \\
SAPLMA-Token & 57.40 & 57.58 & -- & -- \\
MLP Probe & 58.85 & 56.20 & -- & -- \\
HalLocalizer & 54.82 & 52.09 & -- & -- \\
BiGRU Probe & 59.06 & 56.70 & -- & -- \\
CPS-Conv1D & 58.64 & 56.02 & -- & -- \\
Lyapunov Probe & 58.75 & 57.02 & -- & -- \\
\rowcolor{ourslavender}
\textbf{OmniProbe (ours)} & 54.97 & 51.46 & 50.58 & 45.37 \\
\midrule
\multicolumn{5}{l}{\textit{VLM-based methods}} \\
Qwen2.5-VL-72B & 50.97 & 46.38 & 44.86 & 37.15 \\
Qwen3.8-27B & 59.27 & 56.26 & 54.25 & 50.81 \\
Qwen3.6-35B-A3B & 45.62 & 40.97 & 39.34 & 28.52 \\
HalluScope-8B & -- & -- & -- & -- \\
\rowcolor{ourslavender}
\textbf{OmniDetect-27B (ours)} & 69.97 & 70.24 & 63.67 & 65.34 \\
\bottomrule
\end{tabular}
\vspace{3pt}
\begin{minipage}{\textwidth}\footnotesize
Per-VLM probes use each generator's labels; OmniProbe uses only LLaVA labels.
--: no reported subgroup result (Qwen3.5 is not evaluated for per-VLM probes;
HalluScope subgroup scores are unavailable).
\end{minipage}
\end{table}

OmniDetect attains the highest $F_1$ in all four columns, with similar
performance on real and AI-generated images. OmniProbe and the prompted
VLM baselines score lower on AI-generated images in the five-VLM average.
The two image groups also differ in question composition, so these differences
reflect the benchmark subgroups rather than an isolated effect of image generation.

\clearpage
\subsection{OmniProbe Transfer on HalLoc-Instruct}
\label{sec:supp_halloc_transfer}

To evaluate OmniProbe beyond HalluWord, we conduct all 30 directed
transfers among the six VLMs on HalLoc-Instruct~\cite{park2025halloc}.
Each VLM processes the same dataset-provided answers through teacher
forcing, holding response content and annotations fixed while varying
model representations. Features come from the midpoint of each language decoder.

We follow the original dataset splits, using 80,667 training responses,
10,756 validation responses, and all 10,018 test responses. Each source
independently fits PCA-256 and trains a BiGRU probe, with its checkpoint
and threshold selected on the validation split. For each target, Ridge
maps paired training features into the frozen source space without using
hallucination labels; 10\% of the training pairs select the Ridge coefficient
by feature reconstruction error. The source probe and threshold remain
fixed across its five targets. Word scores average constituent token
probabilities; shared segmentation retains punctuation and symbols,
with labels derived from the original character spans.

\begin{figure}[!ht]
\centering
\includegraphics[width=\textwidth]{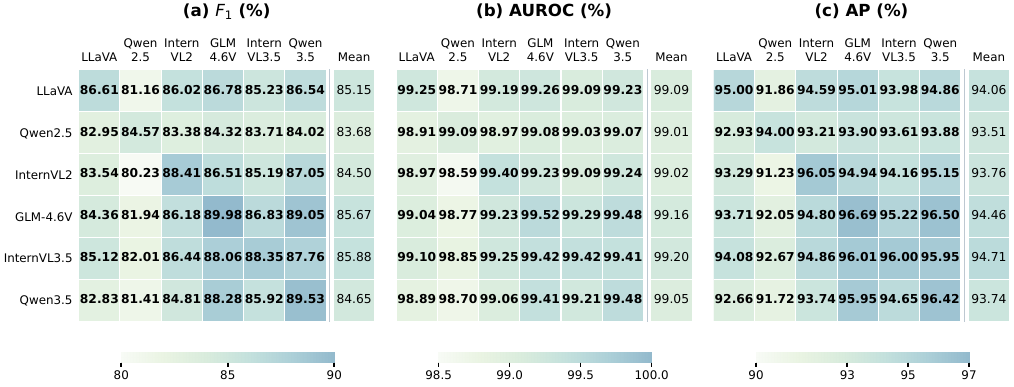}
\caption{\textbf{OmniProbe cross-model transfer on HalLoc-Instruct.}
Word-level $F_1$, AUROC, and AP (\%); rows are sources and columns are targets.
Diagonal cells denote within-model evaluation and are excluded from Mean.
Each metric uses its own color scale.}
\label{fig:supp_halloc_transfer}
\end{figure}

\Cref{fig:supp_halloc_transfer} shows effective probe reuse for every source,
with mean transfer $F_1$ ranging from 83.68\% to 85.88\% and an overall
mean of 84.92\% across the 30 directions. These results extend cross-model
transfer to another dataset under controlled response content, complementing
HalluWord's evaluation on each VLM's native responses.

\clearpage
\subsection{Effect of PCA Dimensionality}
\label{sec:supp_pca_dimension}

We vary the source PCA dimension $k\in\{64,128,256,512\}$ while keeping
LLaVA-1.5-7B as the source and fixing the training data, extraction layer,
BiGRU hidden widths, and random seed. PCA is fitted on the training data
without whitening. Each dimension uses a separately trained source probe
and Ridge maps with the same alignment data and coefficient-selection rule.
\begin{table}[!ht]
\centering\small
\setlength{\tabcolsep}{10pt}\renewcommand{\arraystretch}{1.15}
\caption{\textbf{PCA-dimension ablation of OmniProbe.} Entries report \textbf{word-level $F_1$} (\%). Columns give dimension $k$; bold denotes the best value in each row. Mean excludes the source model.}
\label{tab:supp_pca_dimension}
\begin{tabular*}{\textwidth}{@{\extracolsep{\fill}}lrrrr@{}}
\toprule
VLM & 64 & 128 & 256 & 512 \\
\midrule
LLaVA-1.5-7B (source) & 67.06 & 68.02 & 67.73 & \textbf{68.30} \\
\midrule
Qwen2.5-VL-3B & 51.45 & 51.63 & \textbf{52.97} & 52.15 \\
InternVL2-4B & 54.28 & 54.18 & \textbf{55.58} & 53.51 \\
GLM-4.6V-Flash & 42.83 & \textbf{44.39} & 42.06 & 37.08 \\
InternVL3.5-4B & 47.52 & \textbf{49.01} & 47.20 & 46.79 \\
Qwen3.5-4B & 47.38 & 47.64 & \textbf{47.81} & 45.67 \\
\midrule
Mean & 48.69 & \textbf{49.37} & 49.12 & 47.04 \\
\bottomrule
\end{tabular*}
\end{table}

\Cref{tab:supp_pca_dimension} shows that increasing the representation
dimension does not consistently improve transfer. The 128-dimensional
setting has the highest mean transfer $F_1$ (49.37\%), while 256 dimensions
achieve 49.12\% and the highest scores on three of the five targets.
Increasing to 512 dimensions improves within-source detection but reduces
mean transfer $F_1$ to 47.04\%. The results support moderate dimensionality
as a practical choice for probe reuse, with 256 dimensions offering a
competitive configuration rather than a uniformly optimal one.

\clearpage
\subsection{Effect of Alignment Data Size}
\label{sec:supp_alignment_amount}

We evaluate how much unlabelled paired data OmniProbe requires on
HalLoc-Instruct, fixing the LLaVA source PCA-256 representation, BiGRU
probe, and validation-selected threshold. We vary the alignment budget
from 100 to 20,000 paired responses sampled from the training split and
evaluate transfer to the other five VLMs on the same test split. Within
each budget, 90\% of responses fit Ridge and 10\% select its coefficient
by feature-reconstruction error, without target hallucination labels.
We repeat sampling with three random seeds, using nested subsets within
each seed, and compare against alignment on the full training split.

\begin{figure}[htbp]
    \centering
    \includegraphics[width=\textwidth]{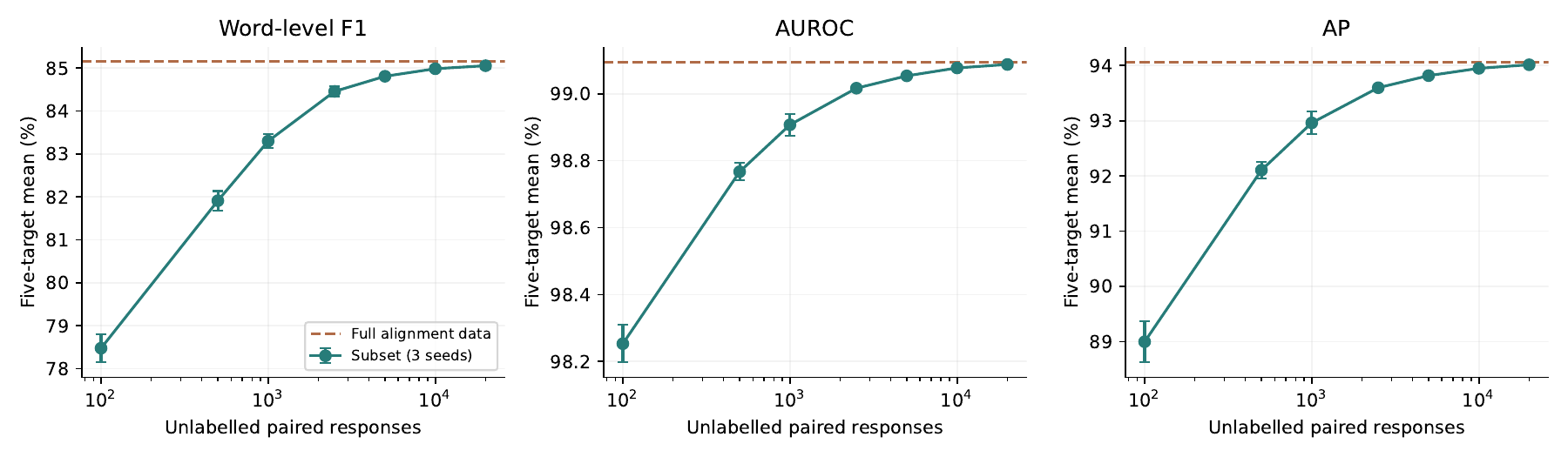}
    \caption{Alignment-data ablation on HalLoc-Instruct. Curves report
    five-target mean scores; error bars show the standard deviation across
    three sampling seeds. Dashed lines denote the full-data alignment
    baseline (80,667 paired responses). The horizontal axis is logarithmic.}
    \label{fig:supp_alignment_amount}
\end{figure}

\begin{table}[!ht]
\centering\small
\setlength{\tabcolsep}{3pt}\renewcommand{\arraystretch}{1.15}
\caption{\textbf{Transfer performance by alignment budget.} Entries report \textbf{word-level $F_1$} (\%). Target scores are averaged over three sampling seeds; Mean gives the five-target average with its standard deviation across seeds. The full-data row reports the single reference run.}
\label{tab:supp_alignment_amount}
\begin{tabular*}{\textwidth}{@{\extracolsep{\fill}}lrrrrrc@{}}
\toprule
\shortstack[l]{Paired\\responses} & \shortstack{Qwen2.5-VL\\3B} & \shortstack{InternVL2\\4B} & \shortstack{GLM-4.6V\\Flash} & \shortstack{InternVL3.5\\4B} & \shortstack{Qwen3.5\\4B} & Mean \\
\midrule
100 & 72.64 & 80.16 & 80.46 & 78.43 & 80.70 & $78.48 \pm 0.33$ \\
500 & 77.18 & 83.22 & 83.85 & 81.49 & 83.81 & $81.91 \pm 0.22$ \\
1,000 & 79.53 & 84.58 & 84.44 & 83.37 & 84.58 & $83.30 \pm 0.16$ \\
2,500 & 80.36 & 85.50 & 85.80 & 84.59 & 86.05 & $84.46 \pm 0.12$ \\
5,000 & 80.81 & 85.70 & 86.39 & 84.87 & 86.28 & $84.81 \pm 0.06$ \\
10,000 & 80.95 & 85.79 & 86.62 & 85.14 & 86.43 & $84.98 \pm 0.05$ \\
20,000 & 81.06 & 85.93 & 86.69 & 85.14 & 86.44 & $85.05 \pm 0.08$ \\
\midrule
80,667 (full) & 81.16 & 86.02 & 86.78 & 85.23 & 86.54 & 85.15 \\
\bottomrule
\end{tabular*}
\end{table}

\Cref{fig:supp_alignment_amount,tab:supp_alignment_amount} show diminishing gains as the alignment
budget increases. With 5,000 paired responses (6.2\% of the full pool),
mean transfer $F_1$ reaches $84.81\pm0.06\%$, only 0.34 percentage points
below full-data alignment (85.15\%). Increasing the budget to 10,000
narrows this gap to 0.16 points. AUROC and AP follow the same trend,
supporting effective probe transfer with a small unlabelled alignment
set while keeping source-detector training unchanged.

\end{document}